\documentclass[10pt, twocolumn, letterpaper]{article}

\newcommand{\final}{1}

\usepackage{cvpr}              

\usepackage{url}            
\usepackage{amsfonts}       
\usepackage{amsmath}
\usepackage{amssymb}
\usepackage{booktabs}       
\usepackage{float}
\usepackage{graphicx}
\usepackage{microtype}      
\usepackage{nicefrac}       
\usepackage{xcolor}         
\usepackage{mathrsfs}
\usepackage{empheq}
\usepackage{array,multirow}

\usepackage[pagebackref,breaklinks,colorlinks]{hyperref}

\makeatletter
\@namedef{ver@everyshi.sty}{}
\makeatother
\usepackage{tikz}

\def\cvprPaperID{2230} 
\def\confName{CVPR}
\def\confYear{2023}

\ifdefined\siggraph
\usepackage{times}
\fi

\usepackage{color}
\usepackage{ifthen}
\usepackage{float}
\usepackage{alltt}
\usepackage{newlfont} 
\usepackage{array}
\usepackage{wrapfig}
\usepackage{booktabs}
\usepackage{multirow}
\usepackage{amsfonts}
\usepackage{dsfont}
\usepackage[linesnumbered,ruled,vlined]{algorithm2e}
 
\SetCommentSty{mycommfont}

\usepackage[capitalize]{cleveref}
\crefname{section}{Sec.}{Secs.}
\Crefname{section}{Section}{Sections}
\Crefname{table}{Table}{Tables}
\crefname{table}{Tab.}{Tabs.}

\usepackage{xcolor}         

\definecolor{DeltaColor}{rgb}{0.039,0.73,0.71}
\definecolor{SetaColor}{rgb}{0.867, 0.0235, 0.376}
\definecolor{SigmaColor}{rgb}{0.98,0.45,0.0}
\definecolor{RedColor}{rgb}{0.8,0,0}
\definecolor{AlphaColor}{rgb}{1.0, 0.4, 0.8}
\definecolor{BetaColor}{rgb}{0.8,0,0.8}
\definecolor{GammaColor}{rgb}{0.5,0,0.7}
\definecolor{EpsilonColor}{rgb}{0.353,0.725,0.906}
\definecolor{TauColor}{rgb}{0.423,0.235,0.192}

\newcommand{\brandon}[1]{{\color{DeltaColor} Brandon: #1 $\blacksquare$}}
\newcommand{\weikai}[1]{{\color{RedColor} Weikai: #1 $\blacksquare$}}
\newcommand{\xiaoxu}[1]{{\color{GammaColor} #1}}
\newcommand{\note}[1]{{\it\color{blue} #1}}

\ifthenelse{\equal{\final}{1}}
{
\renewcommand{\brandon}[1]{}
\renewcommand{\weikai}[1]{}
\renewcommand{\xiaoxu}[1]{}
\renewcommand{\note}[1]{}
}
{}
\newcommand{\modelName}{NeAT}
\newcommand{\netName}{NeAT-Net}
\newcommand{\DFD}{{Deep Fashion 3D Dataset}}
\newcommand{\MGN}{{Multi-Garment Net Dataset}}

\newcommand{\pt}{\mathbf{p}}
\newcommand{\vldty}{\mathcal{V}}
\newcommand{\clr}{c}

\newcommand{\maskPred}{M_{pred}}
\newcommand{\maskGT}{M_{gt}}
\newcommand{\imagePred}{I_{pred}}
\newcommand{\imageGT}{I_{gt}}
\newcommand{\wt}{w(\mathbf{p}(t))}

\begin{document}

\title{\modelName{}: Learning Neural Implicit Surfaces with Arbitrary Topologies\\ from Multi-view Images}
\author{Xiaoxu Meng~~~~~~~~Weikai Chen~~~~~~~~Bo Yang\\
Digital Content Technology Center, Tencent Games\\
{\tt\small \{xiaoxumeng,weikaichen,brandonyang\}@global.tencent.com}
}


\twocolumn[{%
    \renewcommand\twocolumn[1][]{#1}%
    \maketitle
    \begin{center}
    \centering
    \captionsetup{type=figure}
    \includegraphics[width=0.95\textwidth]{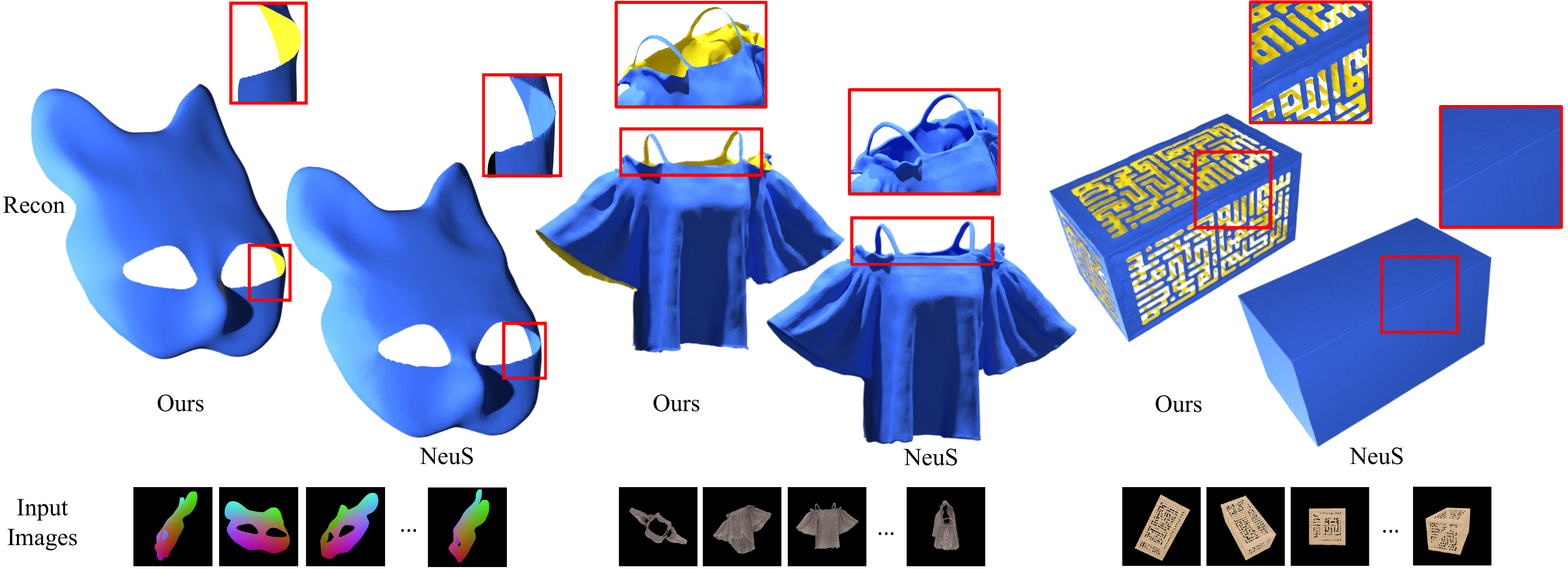}
    \captionof{figure}{
    We show three groups of surface reconstruction from multi-view images. The front and back faces are rendered in blue and yellow respectively. Our method (left) is able to  reconstruct high-fidelity and intricate surfaces of arbitrary topologies, including those non-watertight structures, e.g. the thin single-layer shoulder strap of the top (middle). In comparison, the state-of-the-art NeuS~\cite{wang2021neus} method (right) can only generate watertight surfaces, resulting in thick, double-layer geometries.
    }
    \end{center}%
}]

\begin{abstract}
\vspace{-1em}
Recent progress in neural implicit functions has set new state-of-the-art in reconstructing high-fidelity 3D shapes from a collection of images. 
However, these approaches are limited to closed surfaces as they require the surface to be represented by a signed distance field.
In this paper, we propose \modelName{}, a new neural rendering framework that can learn implicit surfaces with arbitrary topologies from multi-view images. 
In particular, \modelName{} represents the 3D surface as a level set of a signed distance function (SDF) with a validity branch for estimating the surface existence probability at the query positions.
We also develop a novel neural volume rendering method, which uses SDF and validity to calculate the volume opacity and avoids rendering points with low validity. 
\modelName{} supports easy field-to-mesh conversion using the classic Marching Cubes algorithm. 
Extensive experiments on  DTU~\cite{dtu}, MGN~\cite{bhatnagar2019mgn}, and Deep Fashion 3D~\cite{zhu2020deep} datasets indicate that our approach is able to faithfully reconstruct both watertight and non-watertight surfaces.
In particular, \modelName{} significantly outperforms the state-of-the-art methods in the task of open surface reconstruction both quantitatively and qualitatively.
\end{abstract}

\vspace{-1em}
\section{Introduction}
\vspace{-0.5em}

3D reconstruction from multi-view images is a fundamental problem in computer vision and computer graphics. 
Recent advances in neural implicit functions~\cite{yariv2020idr,liu2019learning,wang2021neus,niemeyer2020differentiable} have brought impressive progress in achieving high-fidelity reconstruction of complex geometry even with sparse views. 
They use differentiable rendering to render the inferred implicit surface into images which are compared with the input images for network supervision.
This provides a promising alternative of learning 3D shapes directly from 2D images without 3D data.
However, existing neural rendering methods represent surfaces as signed distance function (SDF)~\cite{yariv2020idr,dist} or occupancy field~\cite{niemeyer2020differentiable, oechsle2021unisurf}, limiting their output to \textit{closed} surfaces.
This leads to a barrier in reconstructing a large variety of real-world objects with open boundaries, such as 3D garments, walls of a scanned 3D scene, \textit{etc}.
The recently proposed NDF~\cite{chibane2020ndf}, 3PSDF~\cite{chen_2022_3psdf} and GIFS~\cite{Ye_2022_CVPR} introduce new implicit representations supporting 3D geometry with arbitrary topologies, including both closed and open surfaces. 
However, none of these representations are compatible with existing neural rendering frameworks.
Leveraging neural implicit rendering to reconstruct \textit{non-watertight} shapes, i.e., shapes with \textit{open} surfaces, from multi-view images remains a virgin land.

We fill this gap by presenting \textit{\modelName{}}, a \textit{Ne}ural rendering framework that reconstructs surfaces with \textit{A}rbitrary \textit{T}opologies using multi-view supervision.
Unlike previous neural rendering frameworks only using color and SDF predictions, we propose a validity branch to estimate the surface existence probability at the query positions, thus avoiding rendering 3D points with low validity as shown in Figure \ref{fig:surface_representation}. In contrast to 3PSDF \cite{chen_2022_3psdf} and GIFS \cite{Ye_2022_CVPR}, our validity estimation is a differentiable process. It is compatible with the volume rendering framework while maintaining its flexibility in representing arbitrary 3D topologies.
To correctly render both closed and open surfaces, we introduce a sign adjustment scheme to render both sides of surfaces, while maintaining unbiased weights and occlusion-aware properties as previous volume renderers.
In addition, to reconstruct intricate geometry, a specially tailored regularization mechanism is proposed to promote the formation of open surfaces.
By minimizing the difference between the rendered and the ground-truth pixels, we can faithfully reconstruct both the validity and SDF field from the input images. 
At reconstruction time, the predicted validity value along with the SDF value can be readily converted to 3D mesh with the classic field-to-mesh conversion techniques, e.g., the Marching Cubes Algorithm~\cite{marching_cubes}. 

We evaluate \modelName{} in the task of multi-view reconstruction on a large variety of challenging shapes, including both closed and open surfaces. 
\modelName{} can consistently outperform the current state-of-the-art methods both qualitatively and quantitatively.  
We also show that \modelName{} can provide efficient supervision for learning complex shape priors that can be used for reconstructing non-watertight surface only from a single image.
Our contributions can be summarized as:
\begin{itemize}
    \item A neat neural rendering scheme of implicit surface, coded \emph{\modelName}, that introduces a novel validity branch, and, \emph{for the first time}, can faithfully reconstruct surfaces with arbitrary topologies from multi-view images.
    \vspace{-2mm}
    \item A specially tailored learning paradigm for \modelName{} with effective regularization for open surfaces.
    \vspace{-2mm}
    \item \modelName~ sets the new state-of-the-art on multi-view reconstruction on open surfaces across a wide range of benchmarks.
\end{itemize}

\section{Related Work}


\paragraph{3D Geometric Representation}
A 3D surface can be represented explicitly with voxels~\cite{shapenet2015,voxnet,7410471,3dr2n2,li2022voxsurf}, point clouds~\cite{8099747,mandikal20183dlmnet,lin2018learning,FoldingNet,achlioptas2018learning}, and meshes~\cite{pixel2mesh,atlasnet,pixel2mesh++}, or can be represented implicitly with neural implicit functions, which have gained popularity for their continuity and the arbitrary-resolution property. 
Watertight surfaces could be represented by occupancy functions~\cite{occnet, sdf0, pifuSHNMKL19}, signed distance functions~\cite{deepsdf,michalkiewicz2019implicit,disn}, or other signed implicits~\cite{Atzmon_2020_CVPR,genova2018unsupervised}. These approaches are limited to closed surfaces as they require the space to be represented as ``inside" and ``outside."
To lift the limitation, unsigned distance function (UDF)~\cite{Venkatesh_2021_ICCV, chibane2020ndf, venkatesh2020dude} is proposed to represent a much broader class of shapes containing open surfaces. However, the signless property of UDF prevents itself from applying the classic field-to-mesh conversion techniques~\cite{marching_cubes}. Instead, these UDF approaches support exporting open surfaces by applying the Ball-Pivoting algorithm~\cite{BPA}, meshUDF~\cite{guillard2021meshudf}, and Neural Dual Contouring~\cite{chen2022ndc}, which are prone to disconnected surface patches with inconsistent normals and coherence artifacts.
GIFS~\cite{Ye_2022_CVPR} represents non-watertight shapes by encoding whether two points are separated by any surface instead of dividing a 3D space into predefined inside/outside regions.
Three-pole signed distance function (3PSDF) ~\cite{chen_2022_3psdf} introduces the \textit{null} sign in addition to the conventional \textit{in} and \textit{out} labels. 
The \textit{null} sign stops the formation of closed isosurfaces, thus enabling the representation of both watertight- and open-surfaces. 
However, 3PSDF~\cite{chen_2022_3psdf} and GIFS~\cite{Ye_2022_CVPR} model the reconstruction of open surfaces as a classification problem, thus preventing these implicit representations from being differentiable. As a result, these methods do not support differentiable downstream tasks like differentiable rendering.
Inspired by the \textit{null} sign of 3PSDF, we propose to represent an open surface as a signed distance function with a validity branch to estimate the surface existence probability at the query positions, which bypasses its limitation of non-differentiability while keeping its capability of modeling arbitrary shapes.

\vspace{-1em}
\paragraph{Implicit Surfaces Reconstruction from Multi-view Images}
It is well-known that a 3D database is more challenging to acquire than a 2D database. As such, learning shapes from 2D supervision is important and necessary.
Multiple differentiable rendering (DR) techniques have been proposed to circumvent the difficulty of explicit correspondence matching in 3D reconstruction.
Two popular types of DR are differentiable rasterization~\cite{kato2018neural,NEURIPS2019_f5ac21cd,softras,8578972} and differentiable ray casting.
A popular branch of differentiable ray casting is surface rendering~\cite{jiang2020sdfdiff, dvr, yariv2020idr, dist, liu2019learning}, which assumes that the ray's color only relies on the color of the intersection point. Surface rendering methods represent the geometry as an implicit function and learn the surface representation from 2D images via differentiable rendering techniques. 
NeRF~\cite{mildenhall2020nerf} and follow-up volume rendering methods assume that the ray's color relies on all the sampled points along the ray. 
UNISURF~\cite{oechsle2021unisurf} improves the reconstruction quality by shrinking the sample region of volume rendering during optimization. 
{
NeuS~\cite{wang2021neus}, VolSDF~\cite{yariv2021volume} and HF-NeuS~\cite{wang2022hfneus} develop volume density functions for watertight surfaces applied to SDFs, which combine the advantages of surface rendering-based and volume rendering-based methods.
However, as the above methods all rely on the representation of the signed distance function or occupancy field, they can only reconstruct watertight shapes. NeuralUDF~\cite{long2023neuraludf} and NeUDF~\cite{liu2023neudf}, two concurrent works, propose to represent arbitrary surfaces as a UDF and develop unbiased density functions that correlate the property of UDF with the volume rendering scheme. However, converting UDF to a mesh typically suffer from artifacts, inconsistent normals, and large computational costs. Compared with the UDF-based approaches, our method represents the scene by a signed distance function with a validity branch,
and thus is compatible with easy field-to-mesh conversion methods, such as the classic Marching Cubes algorithm, ensuring high-fidelity and normal-consistent meshing results from the implicit field.
}

\section{Volumetric Rendering with \modelName}
\begin{figure}[t!]
\centering
    \includegraphics[width=0.9\linewidth]{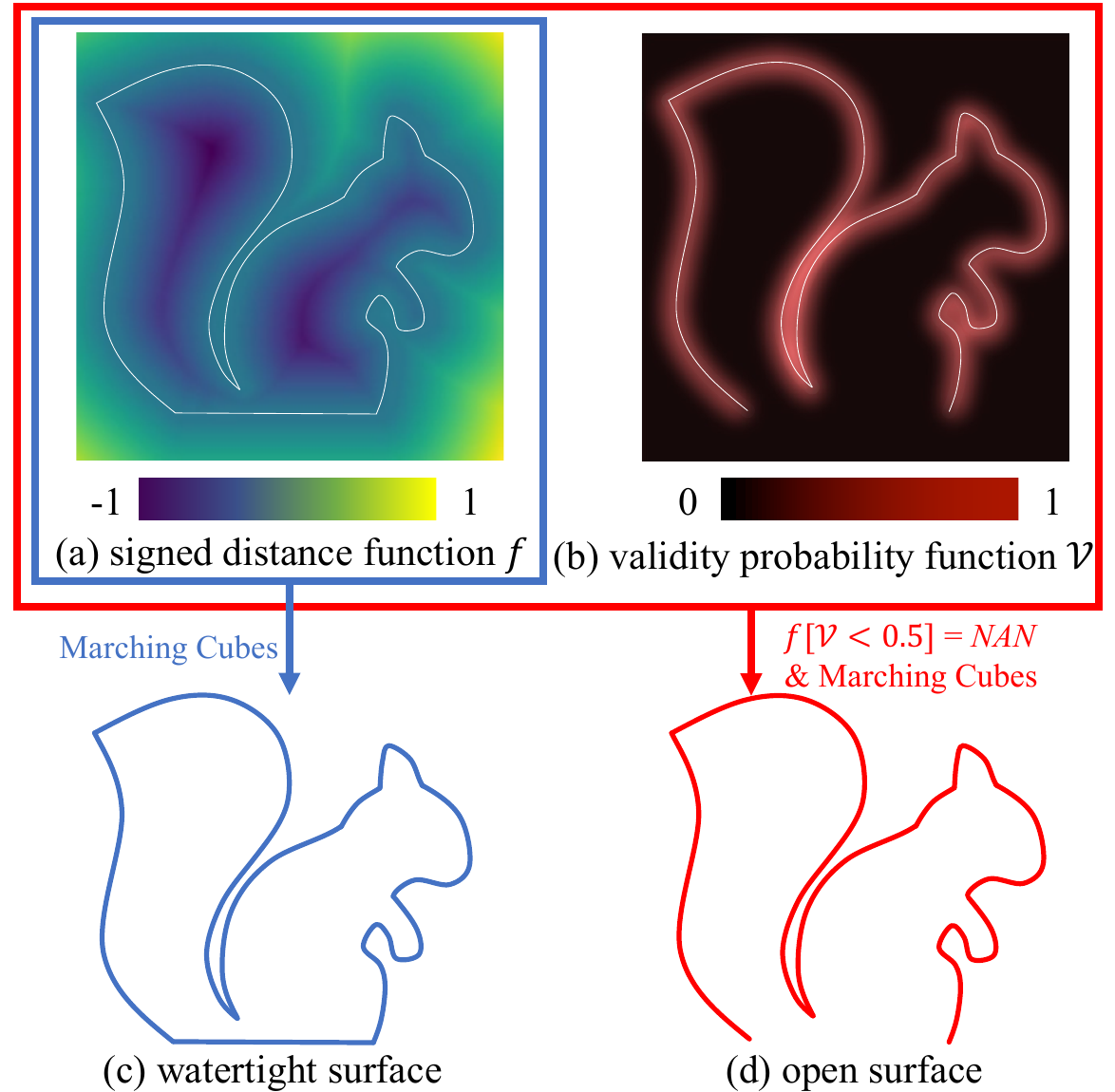}
\caption{(a) is the signed distance function (SDF); (b) is the validity probability function $\vldty$; (c) is the watertight surface extracted from (a) SDF; (d) is the open surface extracted from (a) SDF and (b) validity probability. In our mesh extraction process, we set the SDF of the 3D query points with {low validity (here $\vldty < 0.5$)} to \textit{NAN} and extract the open surface with the Marching Cubes algorithm.
}
\vspace{-1.5em}
\label{fig:surface_representation}
\end{figure}

\begin{figure*}[htbp]
\centering
    \includegraphics[width=0.9\linewidth]{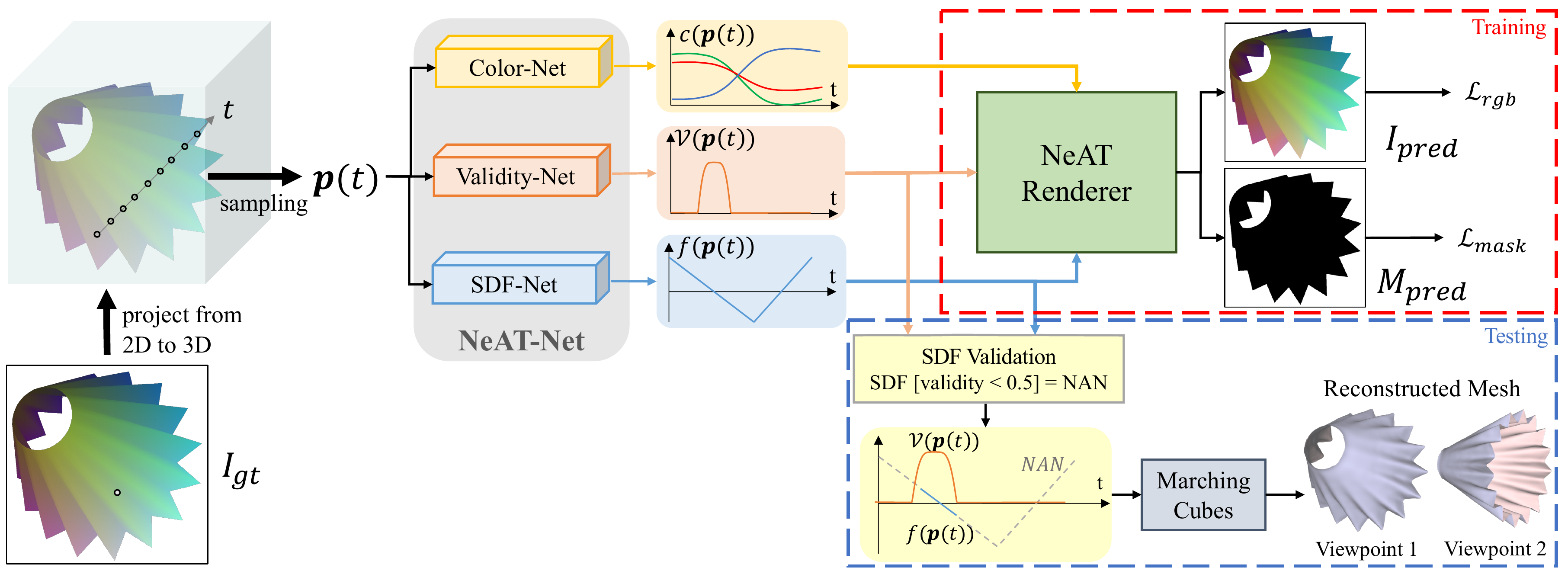}
\vspace{-0.5em}
\caption{The framework of our approach. We project a sampled pixel on the input image $\imageGT$ to 3D to get the sampled 3D points $\textbf{p}(t)$ on a ray. Next, the SDF-Net, Validity-Net, and Color-Net take $\textbf{p}(t)$ as the input to predict the signed distance $f(\textbf{p}(t))$, validity probability $\vldty(\textbf{p}(t))$, and the RGB $\clr(\textbf{p}(t))$, respectively. Then our \modelName~ renderer generates the color $\imagePred$ and the mask $\maskPred$ for \netName~ optimization.
In the mesh exportation (testing) stage, we update the SDF by assigning the low-validity points with a nan value, thus preventing the decision boundary from forming at those regions. Finally, we export arbitrary surfaces from the updated SDF with the Marching Cubes Algorithm.
}
\vspace{-0.5em}
\label{fig:pipeline}
\end{figure*}
Our representation is able to reconstruct 3D surfaces with arbitrary topologies without 3D ground-truth data for training. As shown in Figure~\ref{fig:surface_representation}, by taking the SDF (Figure~\ref{fig:surface_representation} (a)) and validity probability (Figure~\ref{fig:surface_representation} (b)) into consideration together, we acquire additional information that the bottom line in Figure~\ref{fig:surface_representation} (a) is invalid.
{We discard parts of the reconstructed surface according to the validity score and extract an open surface as shown in Figure~\ref{fig:surface_representation} (d) with the Marching Cubes algorithm~\cite{marching_cubes}. }
\subsection{Formulation}
\label{sec:method_overview}
Given N images ${\imageGT (k)}_{k=1}^N$ with a resolution of $(W,\ H)$ together with corresponding camera intrinsics, extrinsics, and object masks ${\maskGT (k)}_{k=1}^N$, our goal is to reconstruct the surface of the object. 
The framework of our method is shown in Figure~\ref{fig:pipeline}. Given a sampled pixel $\mathbf{o}$ on an input image, we project it to the 3D space and get the sampled 3D points on the ray emitting from the pixel as 
$\left\{\mathbf{p}\left(t\right)=\mathbf{o}+t\mathbf{v}\ \middle|\ t\geq0\right\}$,
where $\mathbf{o}$ is the center of the camera and $\mathbf{v}$ is the unit direction vector of the ray. 
Then, we predict the signed distance value $f(\mathbf{p}(t))$, validity probability $\vldty(\mathbf{p}(t))$, 
and the RGB value $\clr(\mathbf{p}(t))$ of the points by our fully connected neural networks called \netName. Specifically, \netName~includes:
\begin{itemize}
	\item SDF-Net: a mapping function $f(\cdot):\mathbf{R}^3\rightarrow \mathbf{R}$ to represent the signed distance field. 
	\item Validity-Net: a mapping function $\vldty(\cdot):\mathbf{R}^3\rightarrow \mathbf{R}$ to represent the validity probability; 
	\item Color-Net: a mapping function $\clr(\cdot):\mathbf{R}^3\rightarrow \mathbf{R}^3$ to predict the per-point color of the 3D space.
\end{itemize}

The outputs of the three networks are delivered to our novel \modelName~renderer to render images and masks from the implicit representations. Our renderer supports both open and closed surfaces, and therefore it provides the capability of reconstructing arbitrary shapes.

{
The predicted mask $\maskPred$ could be inferred from the rendering weights $\wt$ for each sampling point, and the predicted image $\imagePred$ could be calculated from the RGB $\clr(\pt(t))$ and the rendering weights $\wt$:
\vspace{-1em}
\begin{equation}
    \label{equ:pred_imgs}
    \begin{aligned}
        &\maskPred(\mathbf{o}, \mathbf{v}) = \int_{0}^{+\infty}\wt dt,
        \\
        &\imagePred(\mathbf{o}, \mathbf{v}) = \int_{0}^{+\infty}\wt \clr(\pt(t)))dt.
    \end{aligned}
    \vspace{-0.5em}
\end{equation}
}
The predicted masks and images are used for loss calculation during training, which will be illustrated in Section \ref{sec:method_training}. After training is completed, we go through the testing module as shown in Figure \ref{fig:pipeline}; we set the SDFs to NAN for 3D points with $\vldty(\textbf{p})$ less than 0.5, and feed them to Marching Cubes algorithm to produce the final mesh.
\subsection{Construction of \modelName{} Renderer}
\label{sec:method_rendering_model}
According to Equation \ref{equ:pred_imgs}, one key issue in the rendering process is to find an appropriate weight function $\wt$. We split this task into two steps: 1) building a probability density function to estimate volume density from SDF; 2) estimating the weight function $\wt$ from the volume density and the validity probability.

\begin{figure}[t]
\centering
    \includegraphics[width=0.85\linewidth]{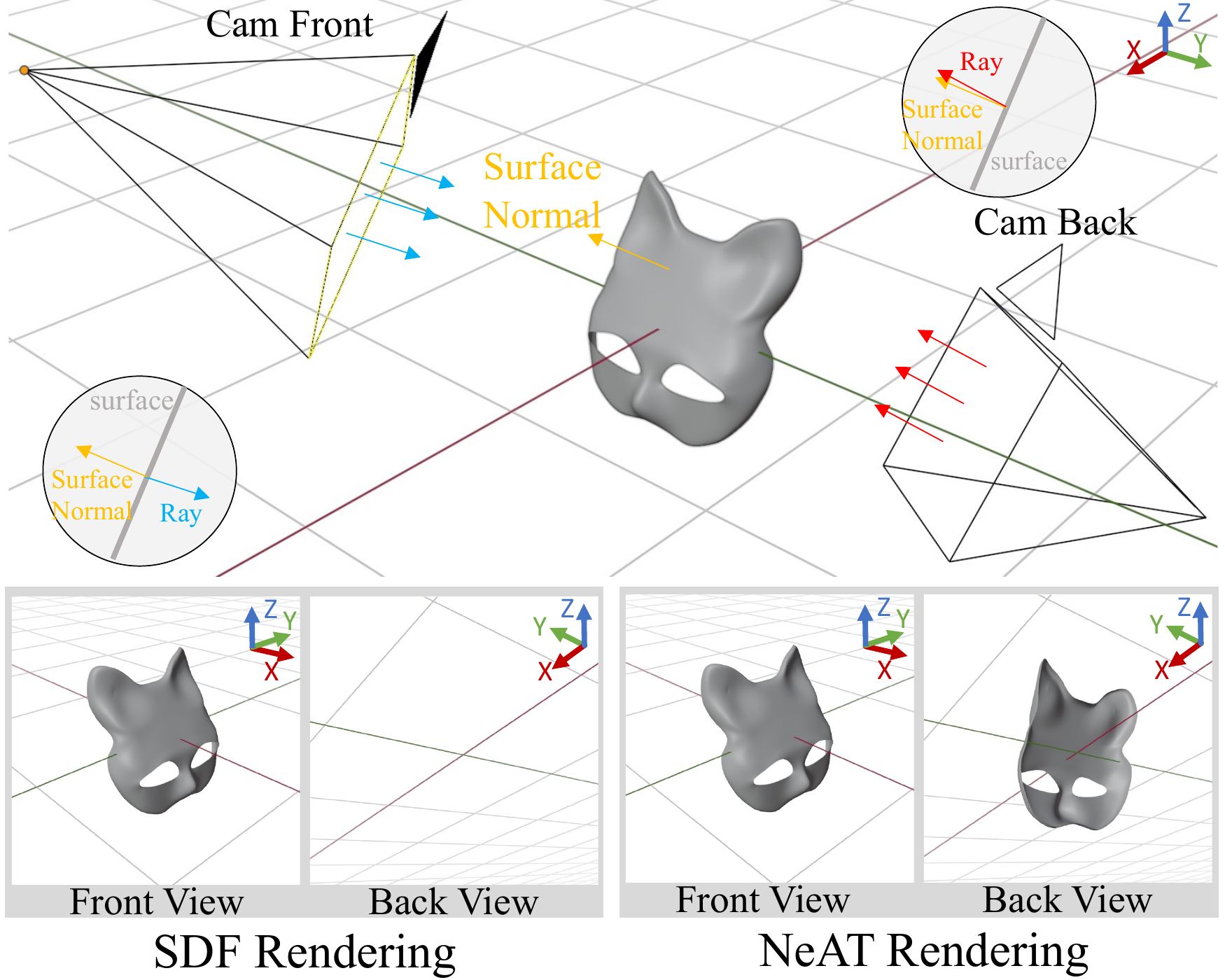}
\caption{
The SDF rendering scheme only renders the surfaces when the ray enters the surface from outside to inside. For an open surface whose surface normal aligns with the back camera's viewing direction, the back camera receives an empty rendering.
Our \modelName~rendering scheme renders both sides of the surfaces.}
\vspace{-1em}
\label{fig:render_bothsides}
\end{figure}

\vspace{-1em}
\paragraph{Construction of Probability Density Function.}
Due to aiming at building arbitrary surfaces, we first introduce the difference between rendering watertight and open surfaces. 
The first difference is the rendering of back-faces. The state-of-the-art watertight surface reconstruction approaches~\cite{wang2021neus,dvr, yariv2020idr} only render the front faces of the surface and ignores the back faces. 
Such a scheme would fail for open surfaces: as shown in Figure~\ref{fig:render_bothsides} (L), the back camera receives an empty rendering of the open surface. While, we render each surface point with ray intersections, as shown in Figure~\ref{fig:render_bothsides} (R).

The second difference is the definition of ``inside" and ``outside", which do not exist for non-watertight surfaces. Therefore, we leverage the local surface normal to determine the sign of the distance as in 3PSDF~\cite{chen_2022_3psdf}. For a local region around a surface, we use positive normal direction as pseudo ``outside" with positive-signed distance, and vice versa. 


We expect that the rendering behaves the same when the ray crosses a surface from either direction. 
The state-of-the-art volume rendering work, NeuS~\cite{wang2021neus}, uses logistic density distribution $\phi_{s}(f(\textbf{p}))$, also known as the derivative of the sigmoid mapping function
${\Phi}_\mathbf{s}\left(f(\textbf{p})\right)$, as the probability density function. 
However, it is not applicable in our scenario -- for surfaces with opposite normal directions, $\Phi_{s}\left(f(\textbf{p})\right)$ will lead to different density values as $\Phi_{s}\left(f(\textbf{p})\right)\neq\Phi_{s}\left(-f(\textbf{p})\right)$.


{
We therefore modify the SDF value by flipping its sign in the regions where the SDF value increases along the camera ray. The probability density function is defined as
\begin{equation}
\sigma(\pt) = \phi_{s}(-Sign(\mathbf{v}\cdot\mathbf{n})f(\pt)),
\vspace{-0.5em}
\end{equation}
}

\noindent where $\mathbf{v}$ is the unit direction vector of the ray and $\mathbf{n}$ is the gradient of the signed distance function. Such definition assures the same rendering behaviors when ray enters the surface from either direction. 

\vspace{-1em}
\paragraph{Construction of Opaque Density Function.}
According to NeuS\cite{wang2021neus}, the weight function $\wt$ should have two properties: unbiased and occlusion-aware. Similarly, we define unbiased rendering weight $\wt$ with Equation~\ref{equ:define_wt_unbiased}~ and define an occlusion-aware weight function based on the opaque density $\rho(t)$ with Equation~\ref{equ:define_wt_occlusion_aware}.

\vspace{-1em}
\begin{empheq}
    [left=\empheqlbrace]{align}
    \wt & = \frac{
            \phi_{s}(-Sign(\mathbf{v}\cdot\mathbf{n})f(\pt(t)))
    }{
    \int_{-\infty}^{+\infty}\phi_{s}(-Sign(\mathbf{v}\cdot\mathbf{n})f(\pt(t)))
    }
    \label{equ:define_wt_unbiased}\\
    \wt &= \exp(-\int_{0}^{t}\rho(u)du)\rho(t)
    \label{equ:define_wt_occlusion_aware}
\end{empheq}




Solving Equation~\ref{equ:define_wt_unbiased}~and Equation~\ref{equ:define_wt_occlusion_aware}, we get
\vspace{-0.5em}
\begin{equation}
    \rho(t) = \frac{
                -\frac{
                    d\Phi_{s}
                }{
                    dt
                }(-Sign(\mathbf{v}\cdot\mathbf{n})f(\pt(t)))
            }{
                \Phi_{s}(-Sign(\mathbf{v}\cdot\mathbf{n})f(\pt(t))))
            }
\vspace{-0.5em}
\end{equation}
\noindent Please checkout the supplemental for the derivation.
\vspace{-1em}
\paragraph{Discretization.}

We adopt the classic discretization scheme in differentiable volumetric rendering\cite{mildenhall2020nerf, wang2021neus} for the opacity and weight function. For a set of sampled points along the ray
$
\{p_{i} = \textbf{o} + t_{i}\textbf{v} |i=1, ..., n, t_{i} < t_{i + 1}\}
$, the rendered pixel color is
\vspace{-1.0em}
\begin{equation}
    \imagePred(\mathbf{o}, \mathbf{v})=\sum_{i=1}^{n}\Pi_{j=1}^{i-1}(1 - \alpha_{j})\alpha_{i}c_{i}
    \label{equ:discrete_color}
    \vspace{-0.5em}
\end{equation}
where $c_i$ is the estimated color for the $i$-th sampling point; $\alpha_{i}$ is the discrete opacity value in SDF rendering
\begin{equation}
\small
\alpha_{i}=\frac{
                \Phi_{s}(-Sign(\mathbf{v}\cdot\mathbf{n})f(\pt(t_{i})))
                 - \Phi_{s}(-Sign(\mathbf{v}\cdot\mathbf{n})f(\pt(t_{i+1})))
            }{
                \Phi_{s}(-Sign(\mathbf{v}\cdot\mathbf{n})f(\pt(t_{i})))
            }
\end{equation}


Now we have built an unbiased and occlusion-aware volume weight function that supports rendering the front and back faces with the SDF representation.

\begin{figure}[t!]
\centering
    \includegraphics[width=0.9\linewidth]{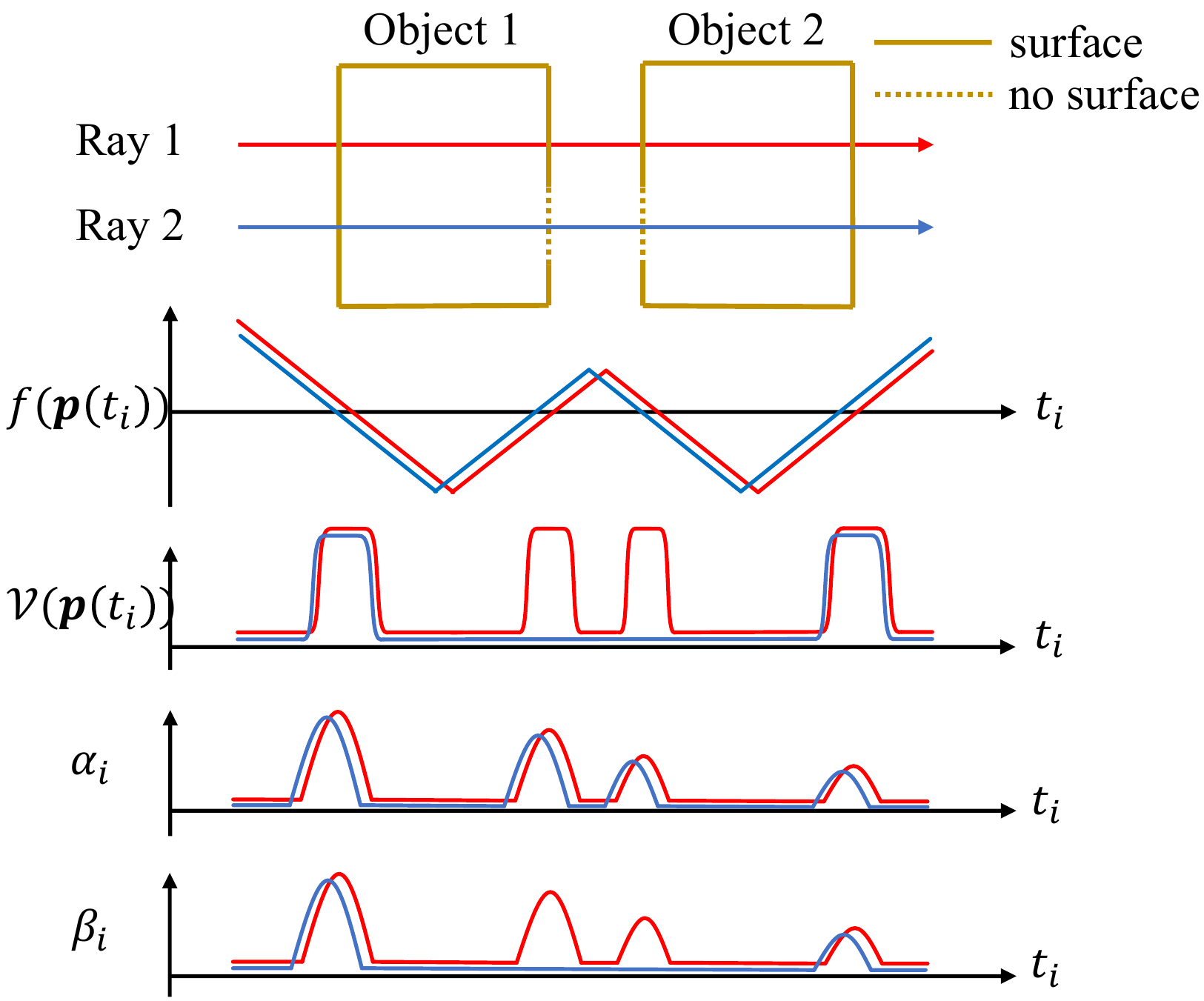}
\vspace{-0.5em}
\caption{Illustration of the rendering with validity probability.}
\vspace{-1em}
\label{fig:render_weights}
\end{figure}

\paragraph{Rendering with Validity Probability.}
To render both closed and open surfaces, we multiply the validity probability of the 3D query points to their opacity value in the rendering process. The discrete opacity value $\beta_{i}$ of the $i$-th sampled point is

\vspace{-0.5em}
\begin{equation}
\beta_{i} = \alpha_{i}\cdot \vldty(\pt(t_{i}))
\end{equation}

We show a 2D illustration of rendering two objects with open boundaries in Figure~\ref{fig:render_weights}.
Ray 2 only has two intersections with the objects due to the existences of open gaps (marked as dotted lines). 
Ray 1 and Ray 2 share the same SDF $f(\pt(t_{i}))$ and discrete opacity value $\alpha_{i}$.
However, according to the validity branch $\vldty(\pt(t_{i}))$, Ray 1 has four valid regions while Ray 2 only has two. By considering the validity probabilities, the discrete opacity value $\beta_{i}$ of the gaps in Ray 2 are set to zero, avoiding generating false surfaces in reconstruction.

Therefore, the final rendered pixel color of a surface is
\begin{equation}
    \imagePred(\mathbf{o}, \mathbf{v})=\sum_{i=1}^{n}\Pi_{j=1}^{i-1}(1 - \beta_{j})\beta_{i}c_{i}
    \vspace{-1em}
    \label{equ:discrete_color_isat}
\end{equation}


\subsection{Training}
\label{sec:method_training}
We supervise the training of \netName{} with five losses. The first three are \textbf{RGB Loss}, \textbf{Mask Loss}, and \textbf{Eikonal Loss}, the same as used in previous neural rendering works~\cite{wang2021neus,yariv2020idr}.
They are defined as
\vspace{-0.5em}
\begin{equation}
    \mathcal{L}_{rgb} = \sum_{i, j}||\imagePred(i,j) - \imageGT(i,j)||\cdot \maskGT(i,j)
    \vspace{-3mm}
\end{equation}
\begin{equation}
    \mathcal{L}_{mask} = \sum_{i, j}BCE(\maskPred(i,j), \maskGT(i,j))
    \vspace{-2mm}
\end{equation}
\begin{equation}
    \mathcal{L}_{eikonal} = \frac{1}{N}\sum_{\textbf{p}}(|\frac{\partial f(\textbf{p})}{\partial \textbf{p}}| - 1)^{2}
    \vspace{-1.5mm}
\end{equation}
where BCE is the binary cross entropy.


\vspace{-0.8em}
\paragraph{Rendering Probability Loss}
In the physical world, the existence of surfaces is binary (exist/not exist). As a result, the validity probability of a 3D sampling point is either 0 (with no surface) or 1 (with surface). We therefore add the binary cross entropy of $\vldty(\textbf{p})$ as an extra regularization:
\vspace{-2mm}
\begin{equation}
    \mathcal{L}_{bce} = \frac{1}{N}\sum_{\textbf{p}} BCE(\vldty(\textbf{p}), \vldty(\textbf{p})).
    \vspace{-3mm}
\end{equation}

\begin{figure*}[htbp]
    \vspace{-6mm}
    \begin{minipage}[t]{.16\textwidth}
        \centering
        \includegraphics[width=0.9\textwidth]{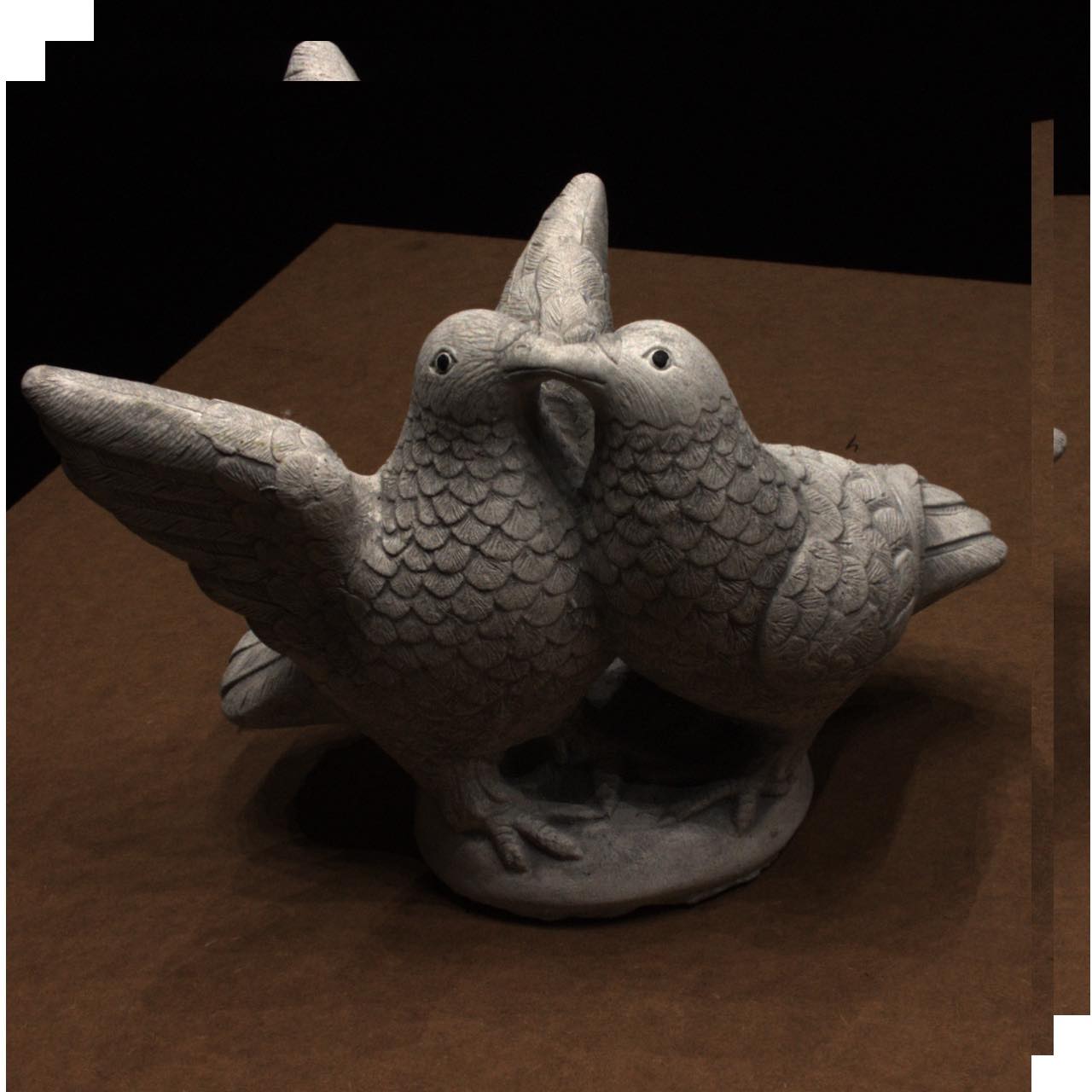}
    \end{minipage}
    \begin{minipage}[t]{.16\textwidth}
        \centering
        \includegraphics[width=\textwidth]{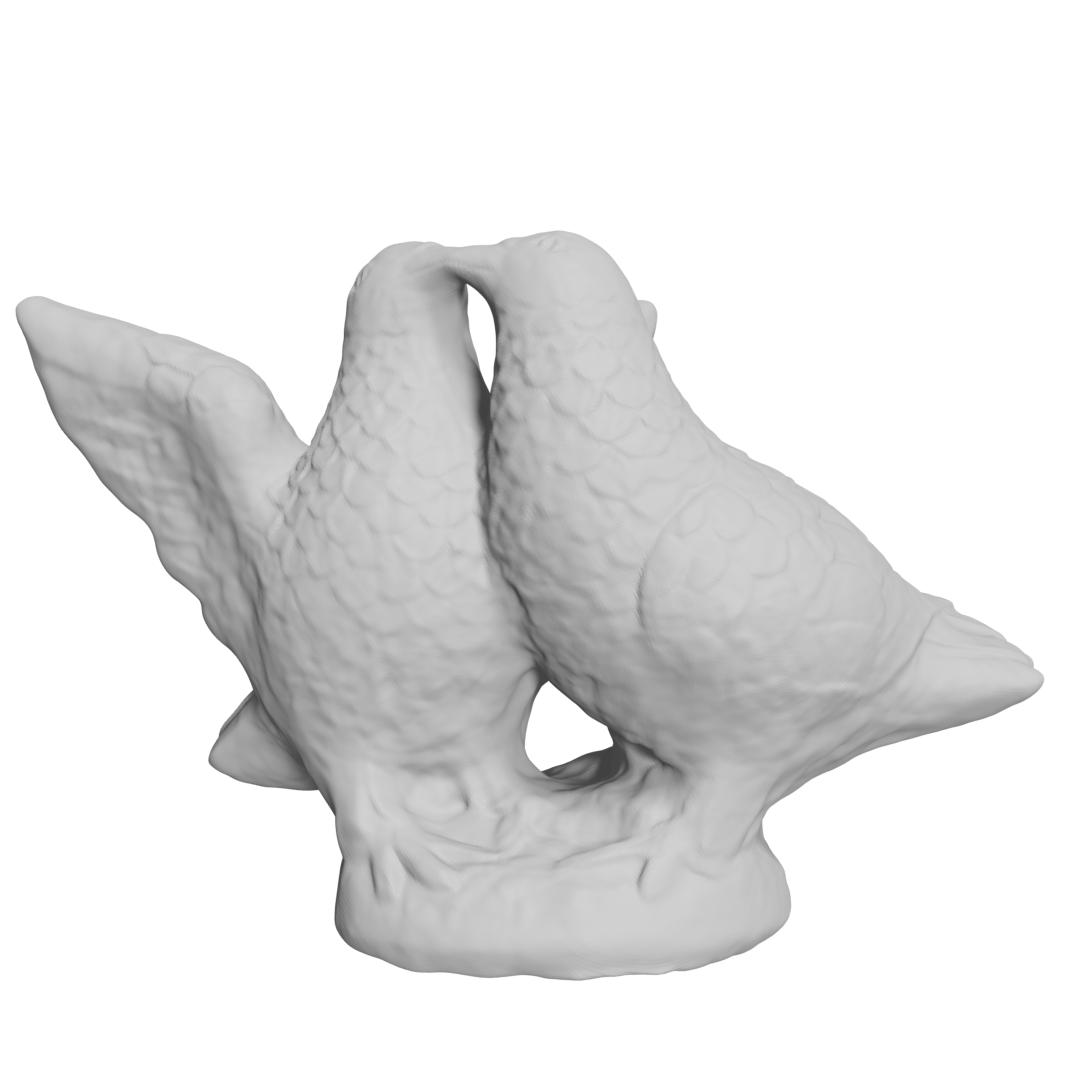}
    \end{minipage}  
    \begin{minipage}[t]{.16\textwidth}
        \centering
        \includegraphics[width=\textwidth]{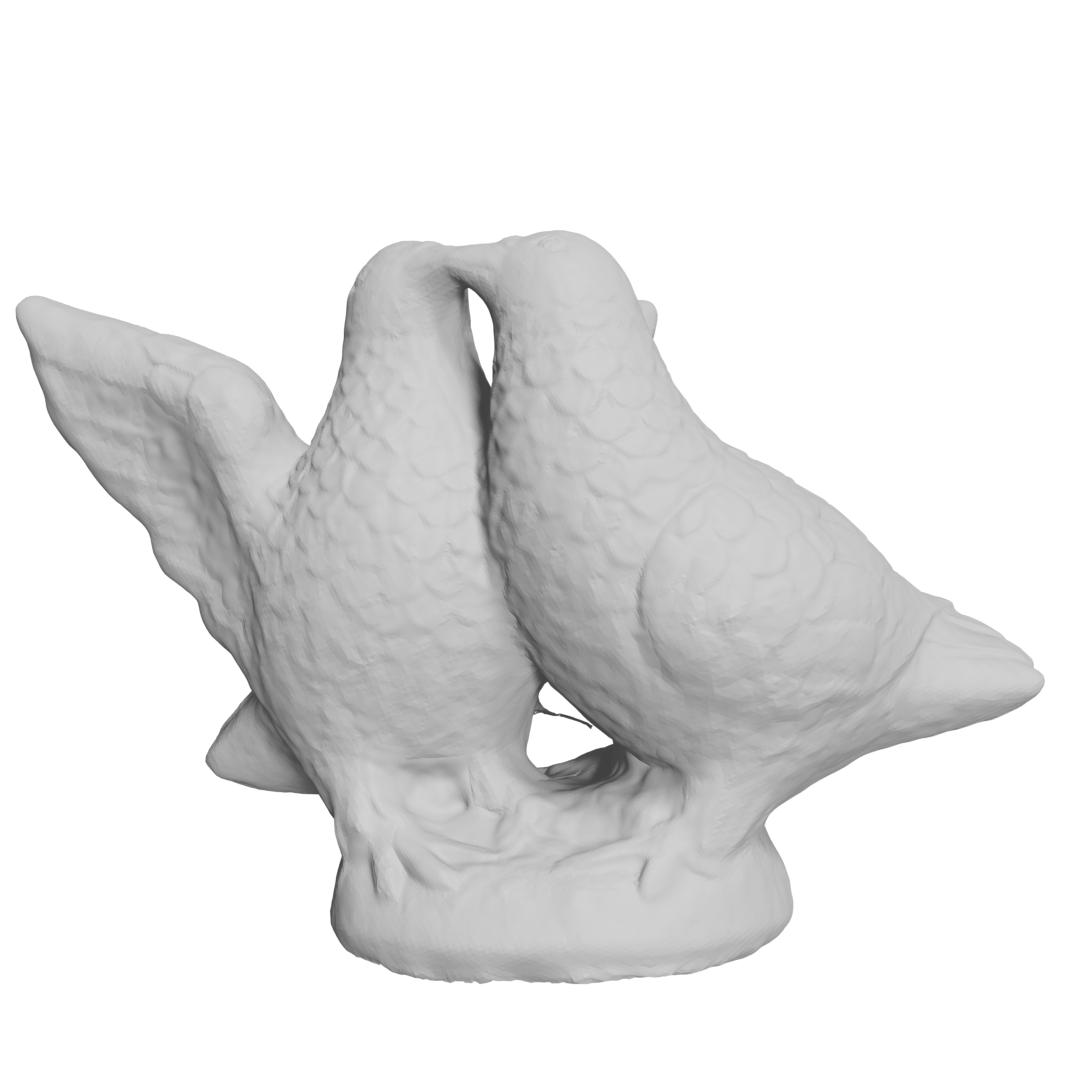}
    \end{minipage}
    \begin{minipage}[t]{.16\textwidth}
        \centering
        \includegraphics[width=\textwidth]{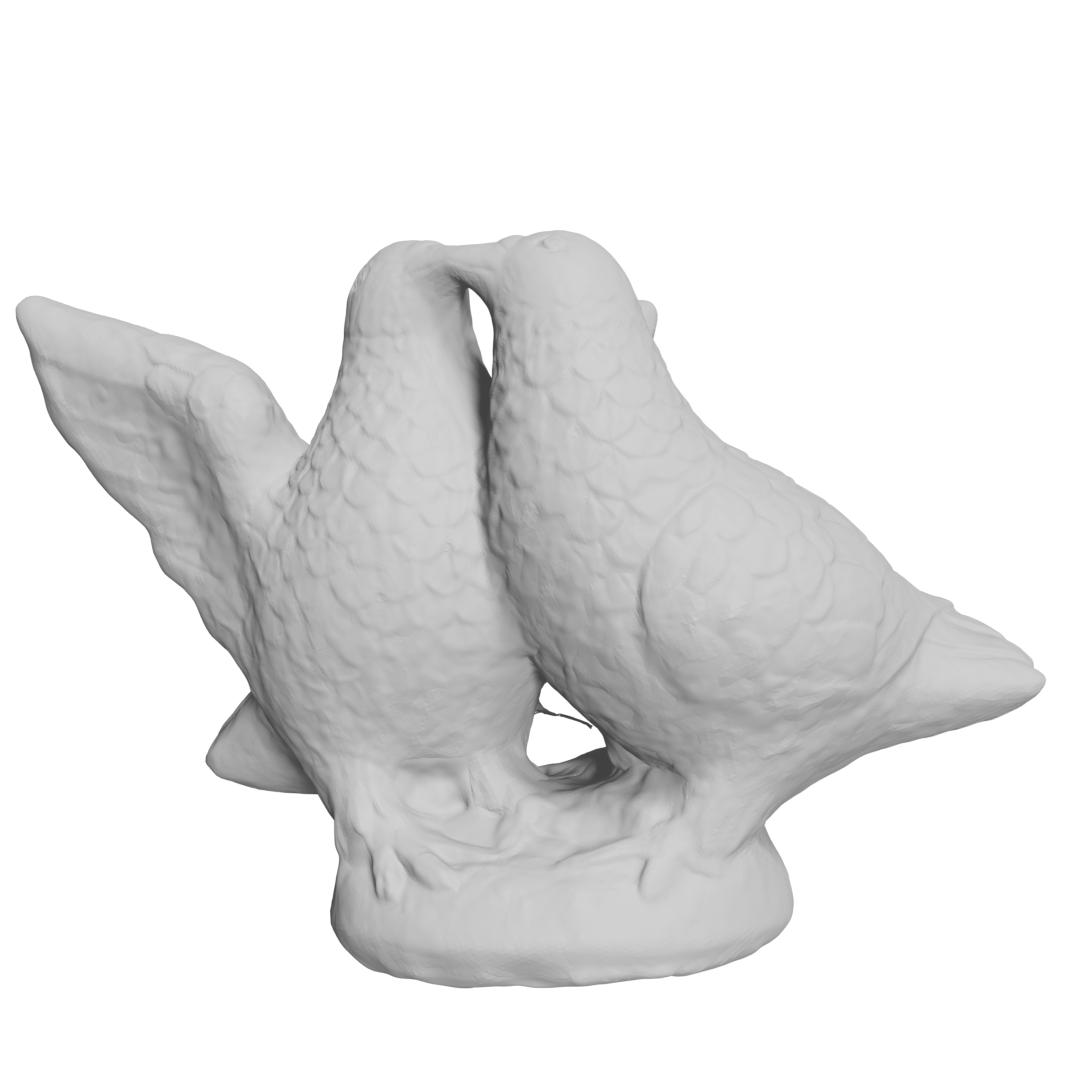}
    \end{minipage}
    \begin{minipage}[t]{.16\textwidth}
        \centering
        \includegraphics[width=\textwidth]{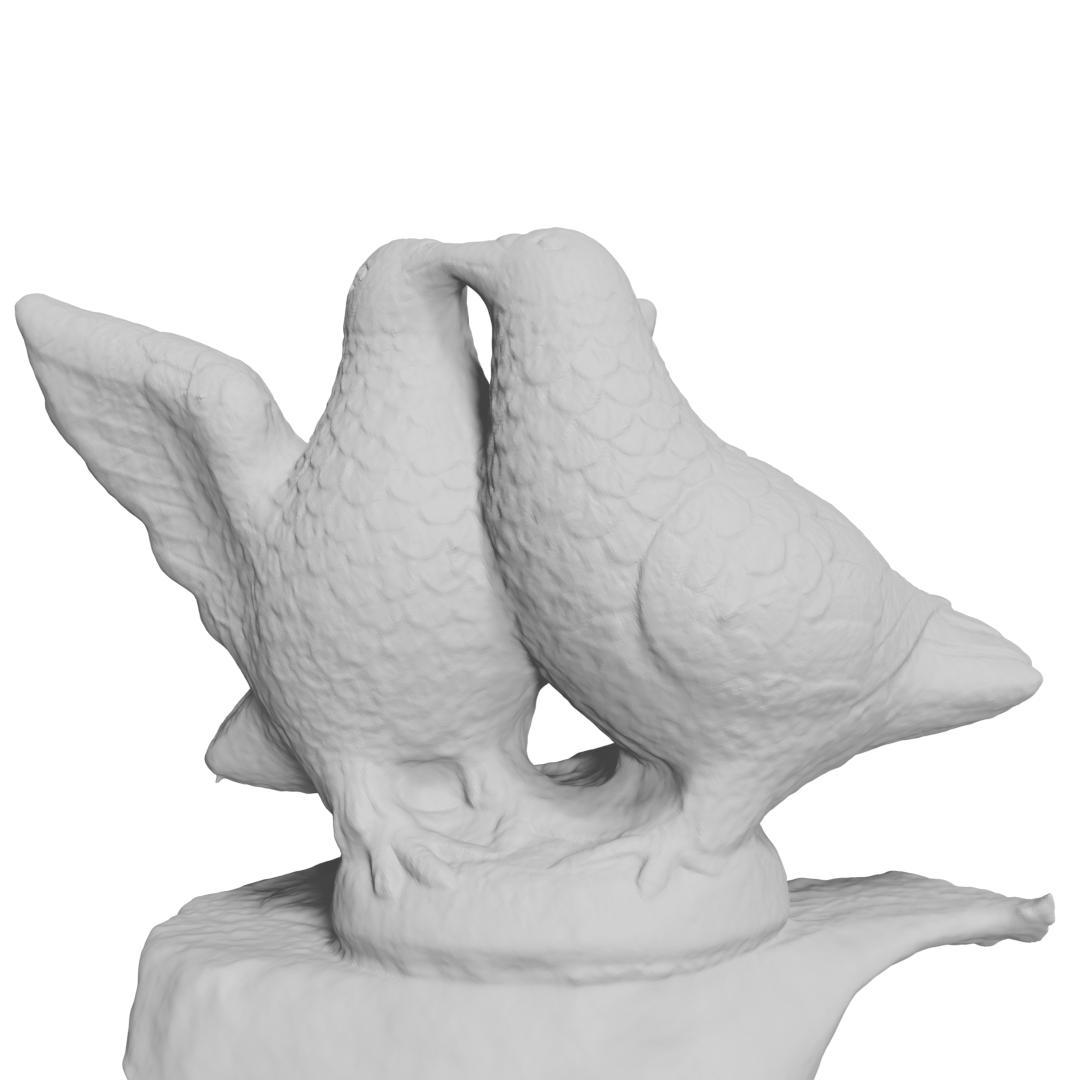}
    \end{minipage}
    \begin{minipage}[t]{.16\textwidth}
        \centering
        \includegraphics[width=\textwidth]{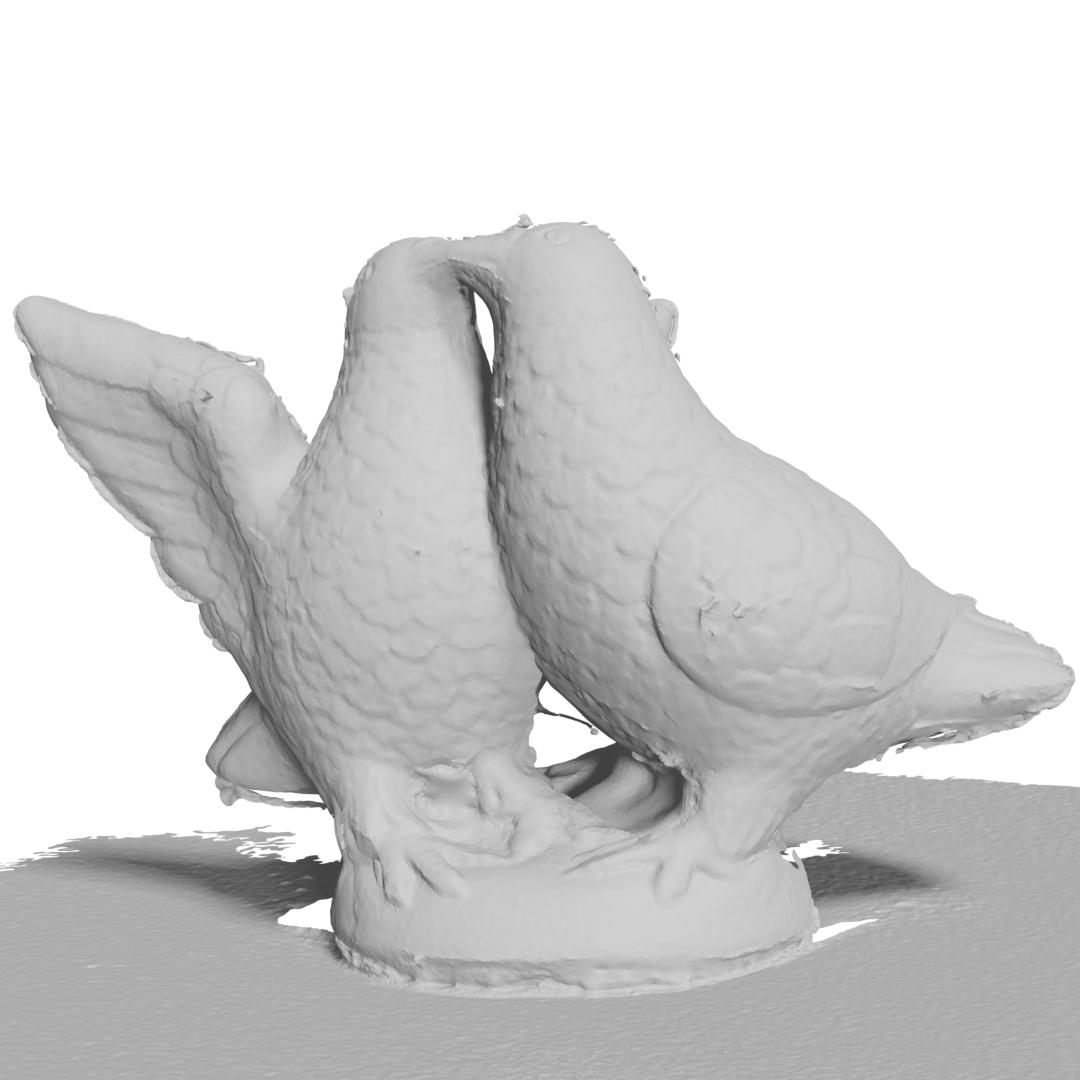}
    \end{minipage}
    \vspace{-0mm}
    \\
    \begin{minipage}[t]{.16\textwidth}
        \centering
        \includegraphics[width=0.9\textwidth]{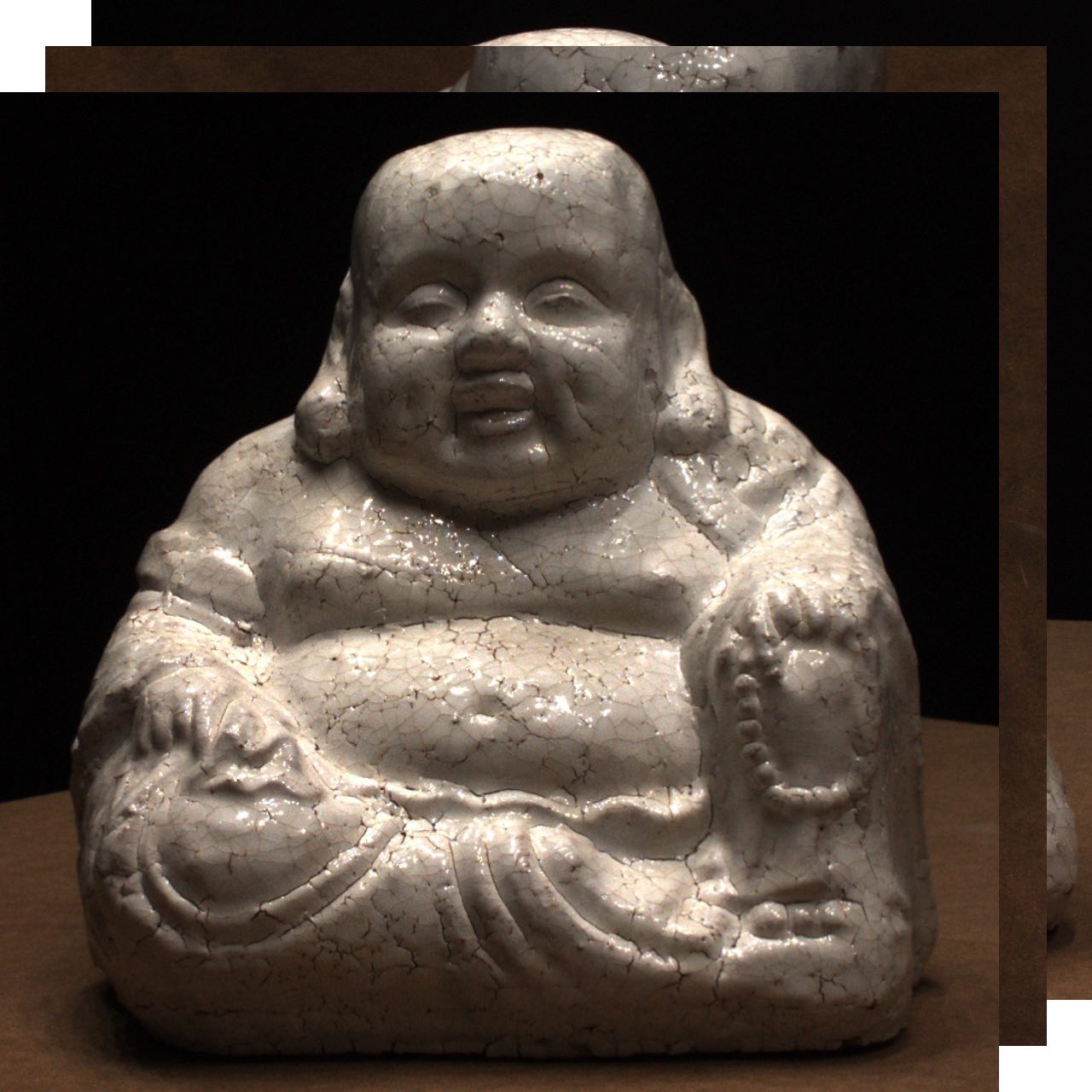}
    \end{minipage}
    \begin{minipage}[t]{.16\textwidth}
        \centering
        \includegraphics[width=\textwidth]{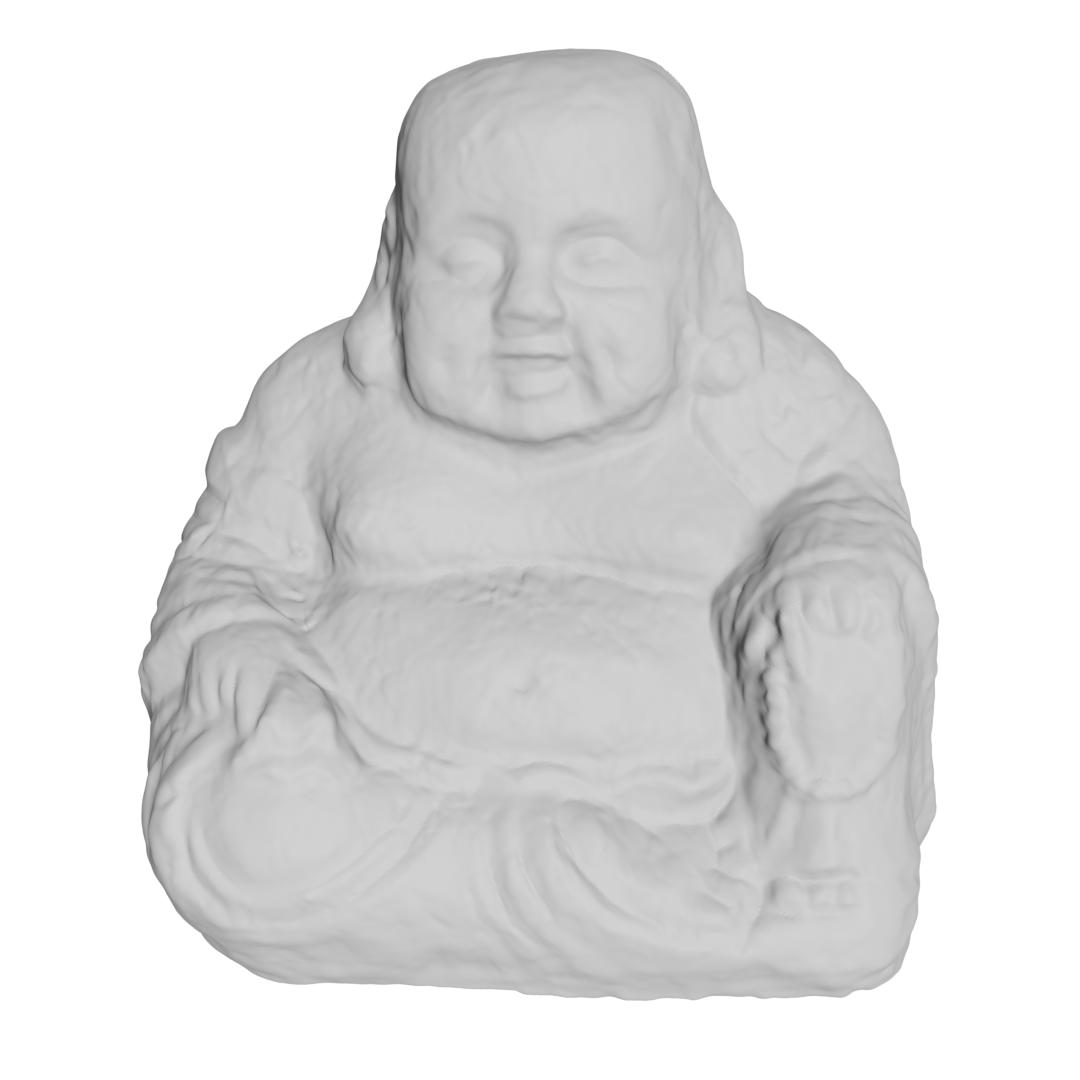}
    \end{minipage}  
    \begin{minipage}[t]{.16\textwidth}
        \centering
        \includegraphics[width=\textwidth]{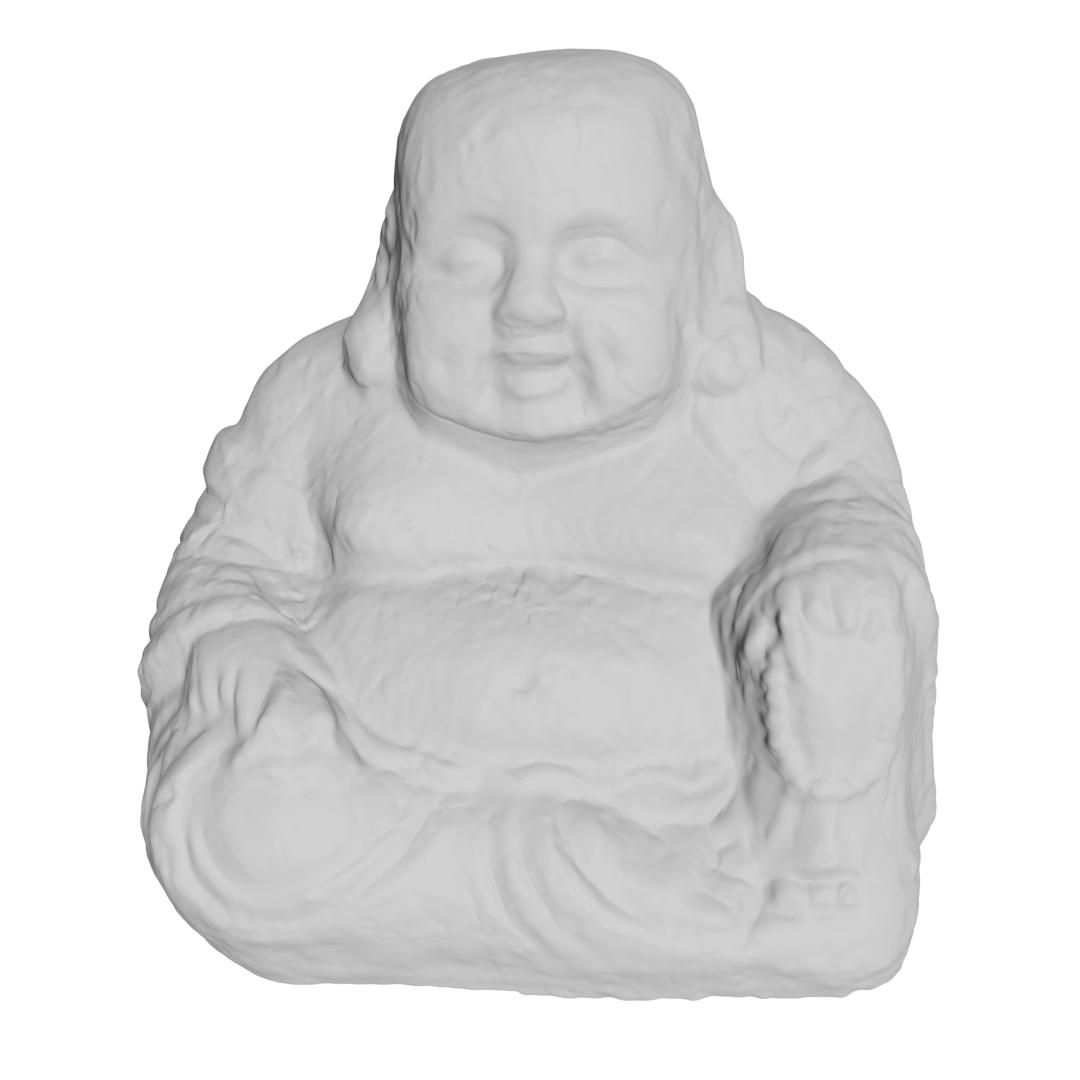}
    \end{minipage}
    \begin{minipage}[t]{.16\textwidth}
        \centering
        \includegraphics[width=\textwidth]{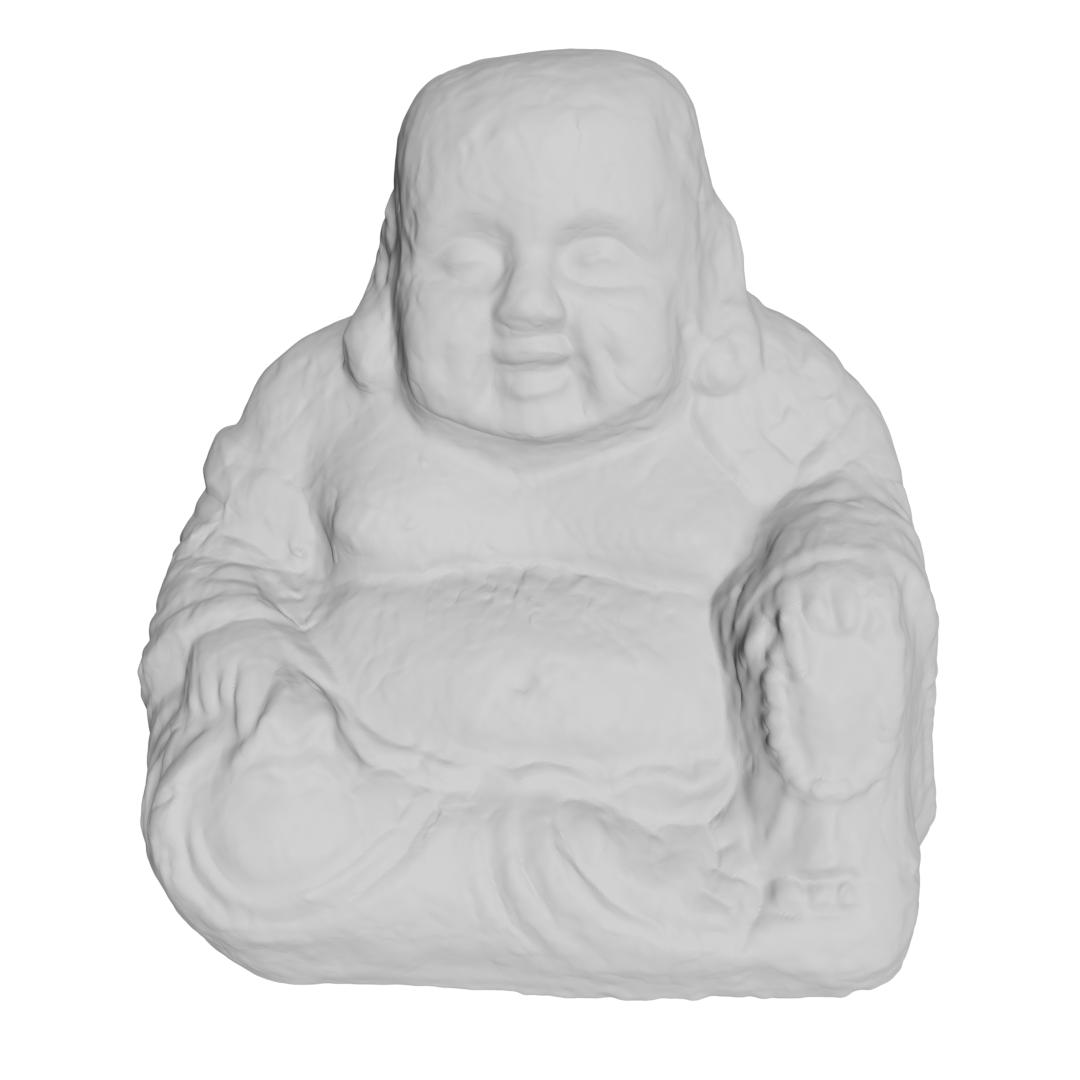}
    \end{minipage}
    \begin{minipage}[t]{.16\textwidth}
        \centering
        \includegraphics[width=\textwidth]{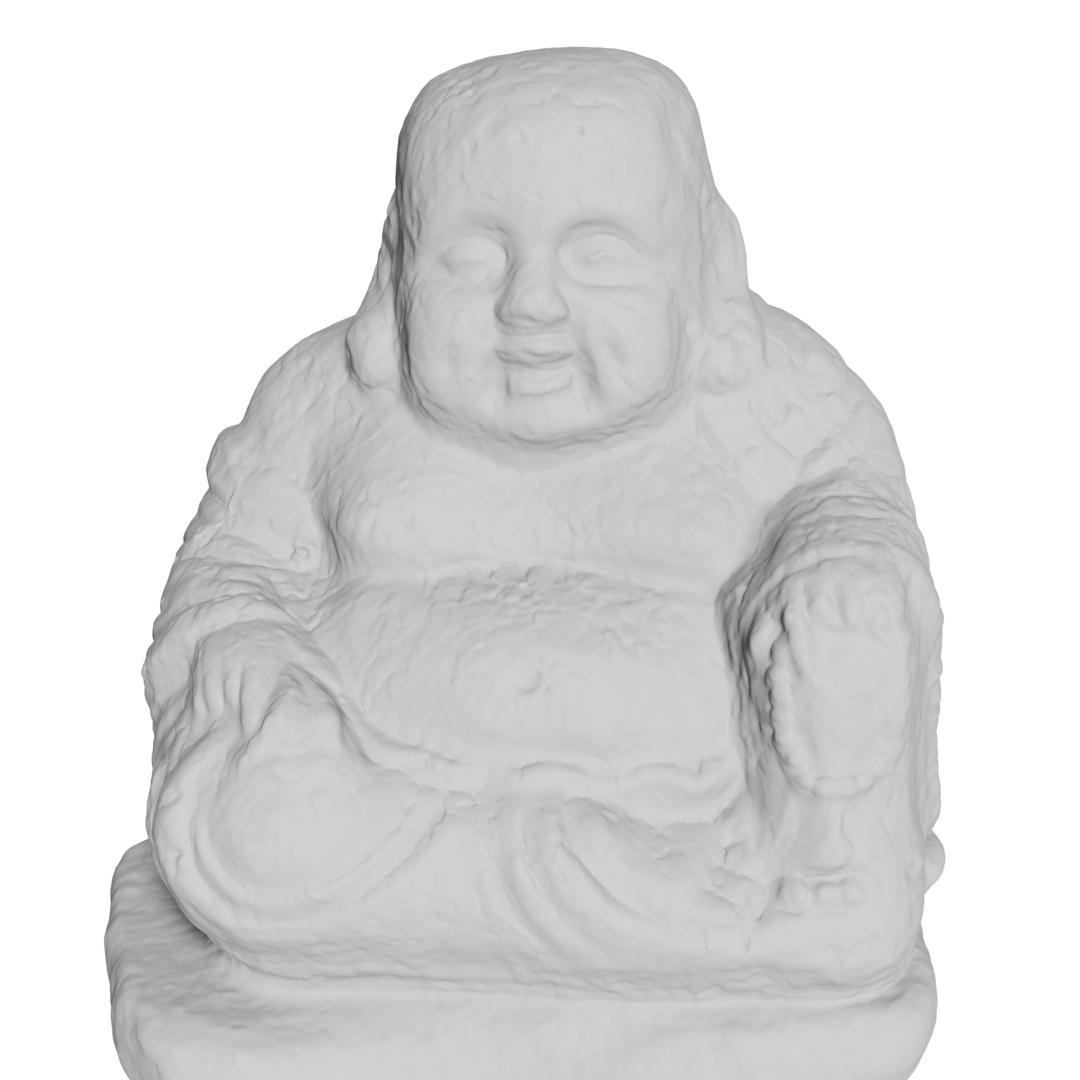}
    \end{minipage}
    \begin{minipage}[t]{.16\textwidth}
        \centering
        \includegraphics[width=\textwidth]{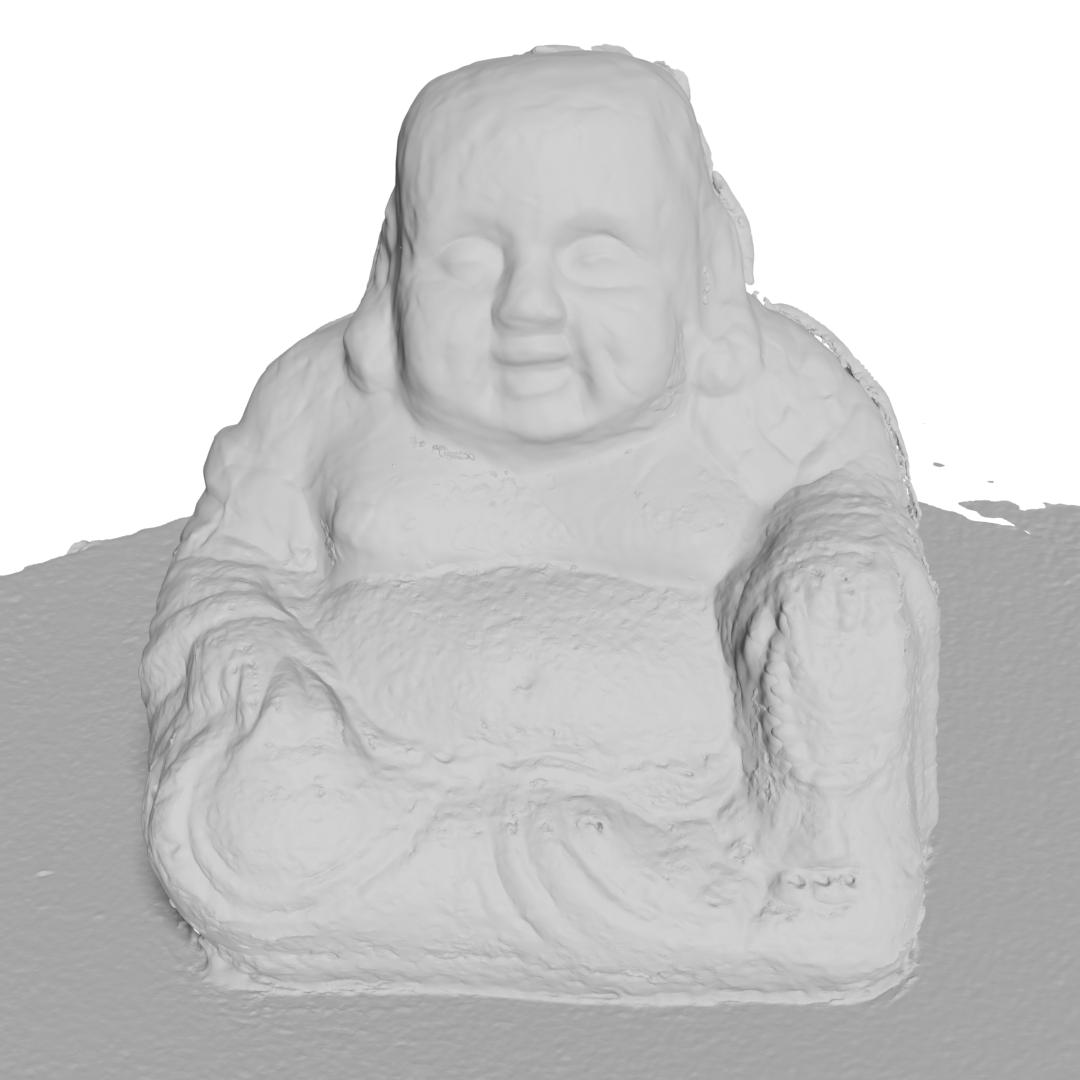}
    \end{minipage}
    \\
    \begin{minipage}[t]{.18\textwidth}
        \centering
        \subfloat[Ref Images]{\includegraphics[width=0.6\textwidth]{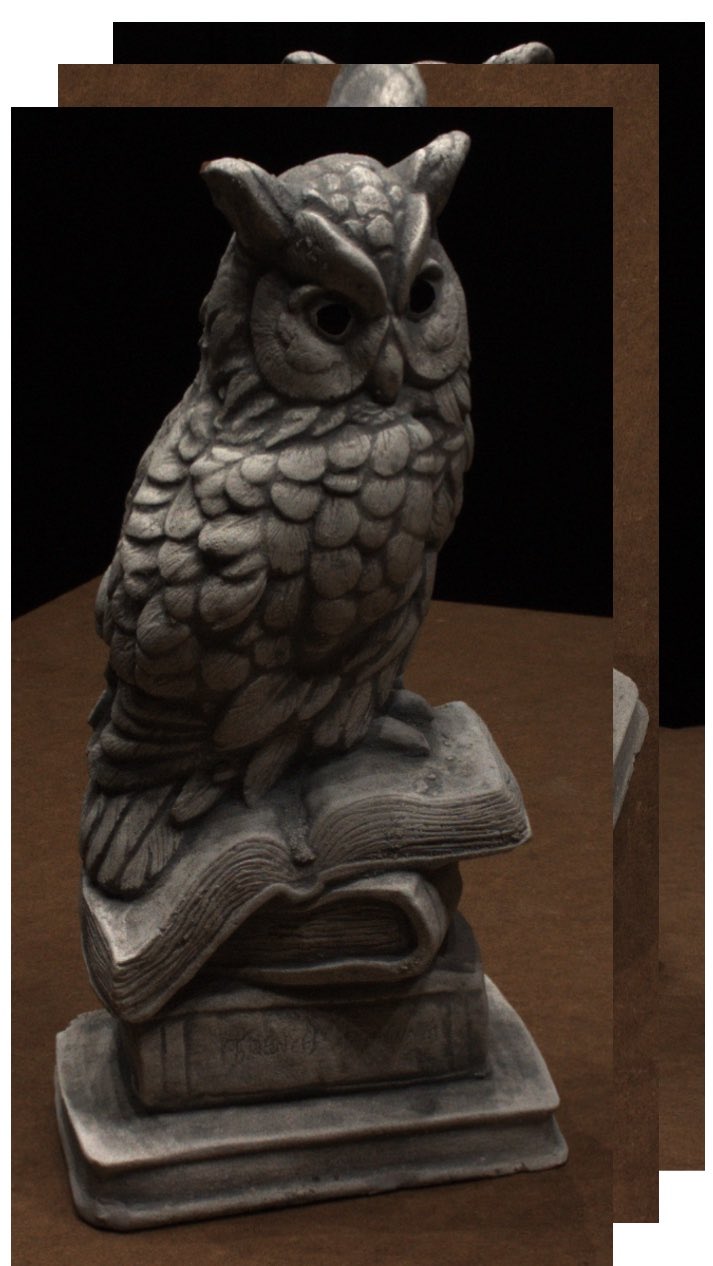}}
    \end{minipage}
    \begin{minipage}[t]{.16\textwidth}
        \centering
        \subfloat[Ours]{\includegraphics[width=0.9\textwidth]{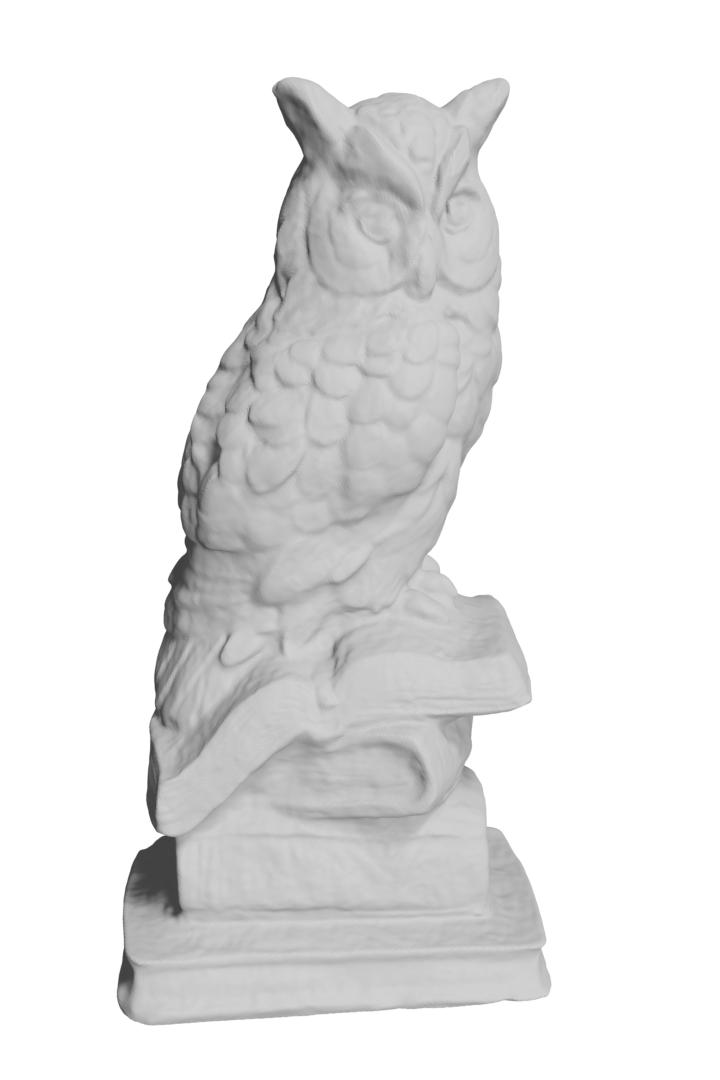}}
    \end{minipage}  
    \begin{minipage}[t]{.16\textwidth}
        \centering
        \subfloat[NeuS]{\includegraphics[width=0.9\textwidth]{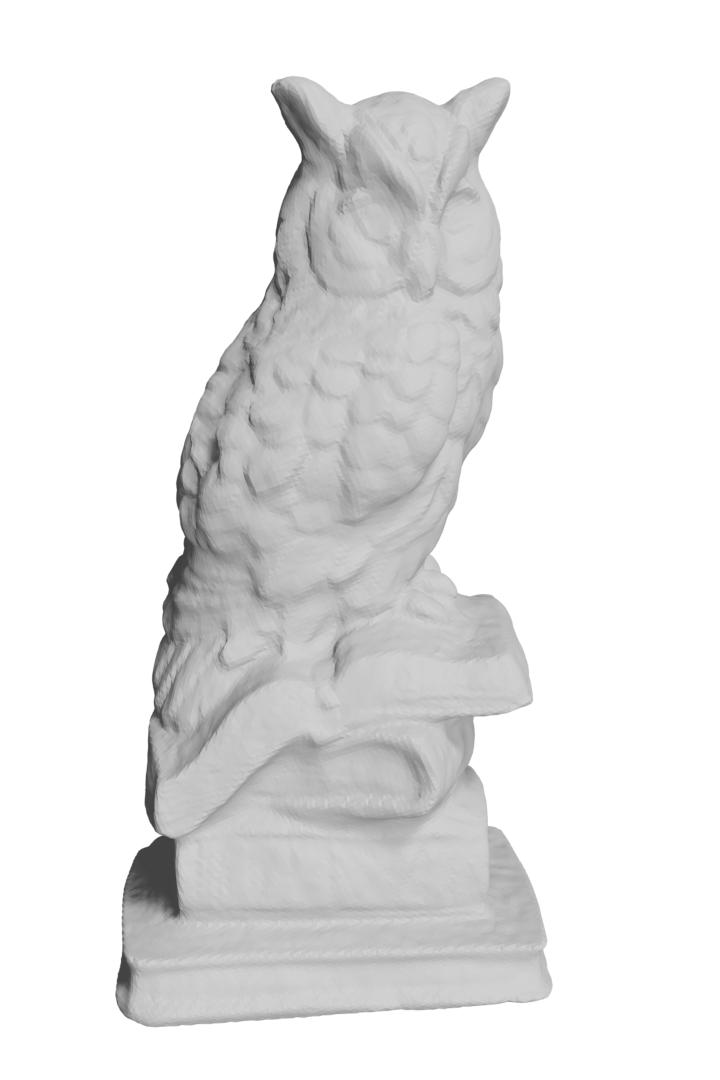}}
    \end{minipage}
    \begin{minipage}[t]{.16\textwidth}
        \centering
        \subfloat[IDR]{\includegraphics[width=0.9\textwidth]{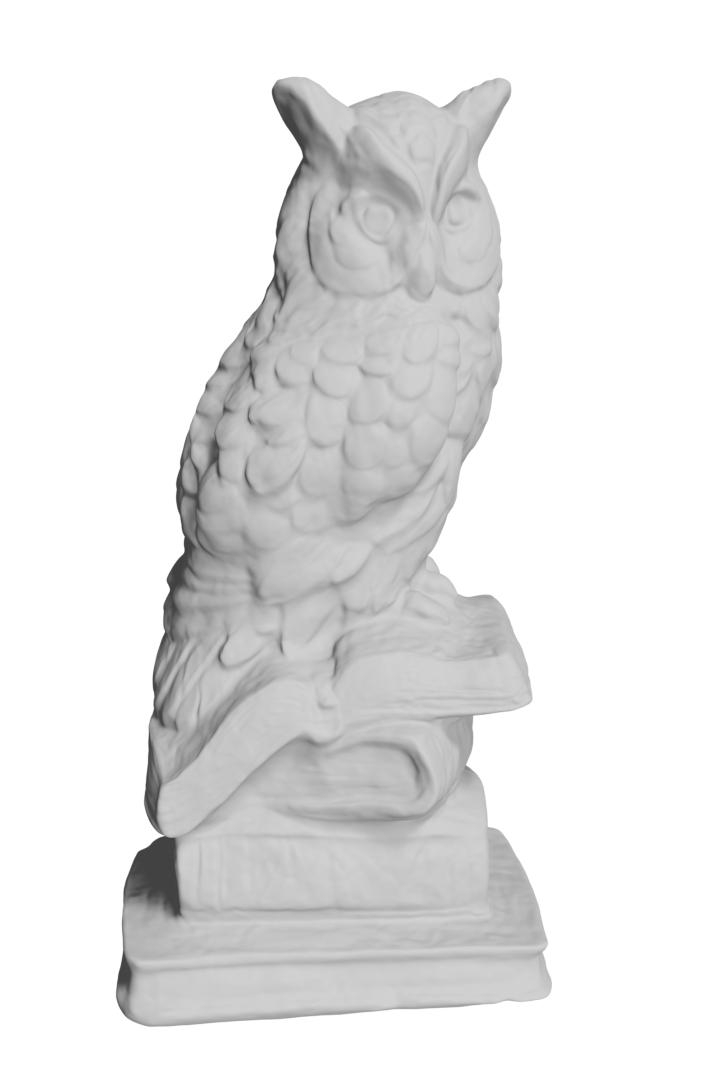}}
    \end{minipage}
    \begin{minipage}[t]{.16\textwidth}
        \centering
        \subfloat[HFS]{\includegraphics[width=0.9\textwidth]{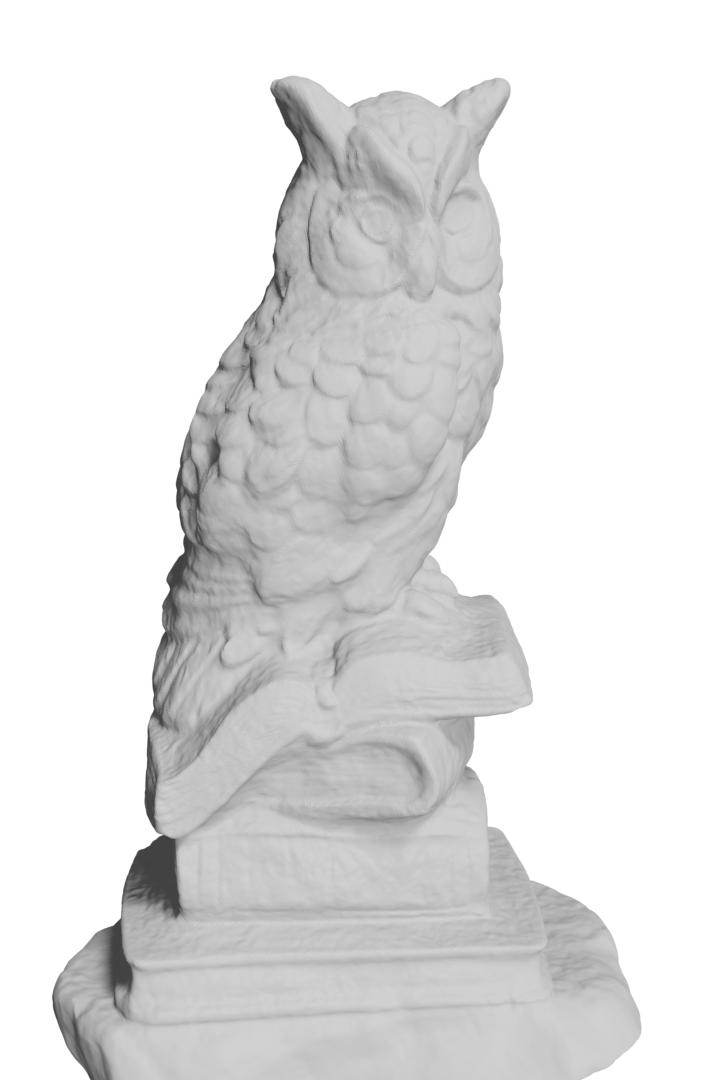}}
    \end{minipage}
    \begin{minipage}[t]{.16\textwidth}
        \centering
        \subfloat[COLMAP]{\includegraphics[width=0.9\textwidth]{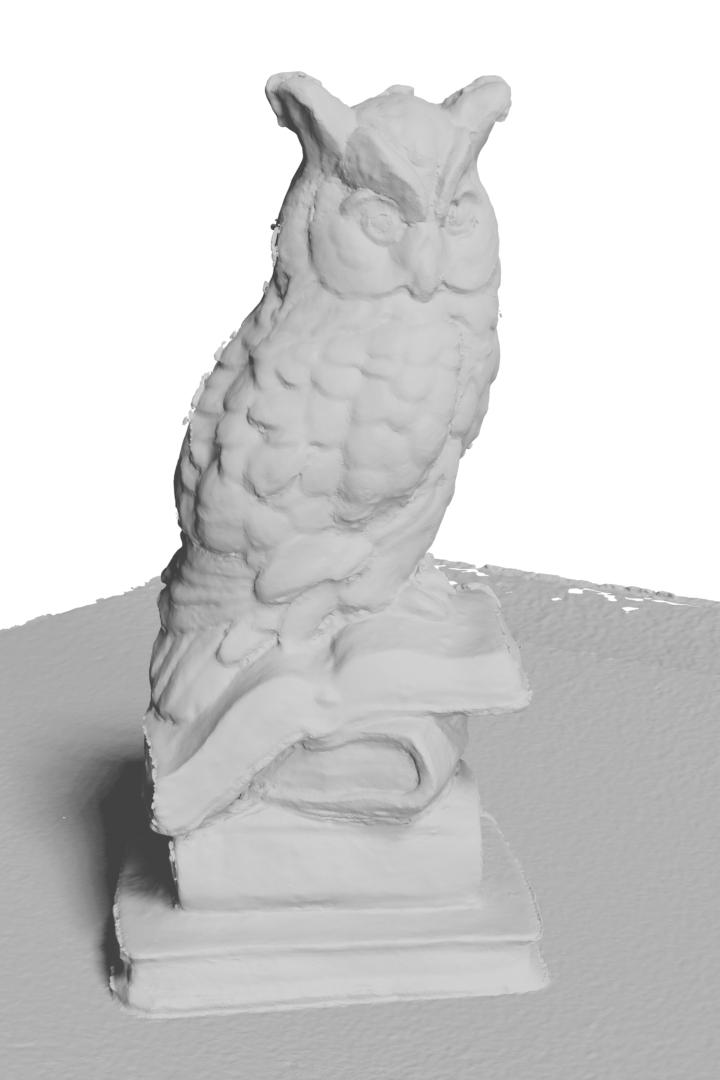}}
    \end{minipage}
\vspace{-0.4em}
\caption{Comparisons on watertight surface reconstruction.}
\vspace{-1.5em}
\label{fig:comparison_watertight}
\end{figure*}

\vspace{-1em}
\paragraph{Rendering Probability Regularization}
For real-world objects with open structures, the surfaces are sparsely distributed in the 3D space. To prevent \netName{} from predicting redundant surfaces, we introduce a sparsity loss to promote the formation of open surfaces:
\vspace{-2mm}
\begin{equation}
    \mathcal{L}_{sparse} = \frac{1}{N}\sum_{\textbf{p}}\vldty(\textbf{p}).
    \vspace{-3mm}
\end{equation}

\noindent We optimize the following loss function
\vspace{-1.5mm}
\begin{align}
\begin{split}
\mathcal{L} &= \mathcal{L}_{rgb}
            + \lambda_{mask} \cdot \mathcal{L}_{mask}
            + \lambda_{eikonal} \cdot \mathcal{L}_{eikonal}\\
            & + \lambda_{bce} \cdot \mathcal{L}_{bce} 
            + \lambda_{sparse} \cdot \mathcal{L}_{sparse}.
\end{split}
\vspace{-1.5mm}
\end{align}

\section{Experiments}
\subsection{Experiment Setup}

\paragraph{Tasks and Datasets.}
We validate \modelName~using three types of experiments. 
We first conduct multi-view reconstruction for real-world watertight objects to ensure that \modelName~achieves comparable reconstruction quality on watertight surfaces. We conduct this experiment on 10 scenes from the \textit{DTU Dataset}~\cite{dtu}. Each scene contains $49$ or $64$ RGB images and masks with a resolution of $1600\times 1200$.
Second, we reconstruct open surfaces from multi-view images. We run this experiment on eight categories from the \DFD~\cite{zhu2020deep} and five categories from the \MGN~\cite{bhatnagar2019mgn}, which contain clothes with a wide variety of materials, appearance, and geometry, including challenging cases for reconstruction algorithms, such as camisoles.
Finally, we construct an autoencoder, which takes a single image as the input and provides validation on the challenging task of single-view reconstruction on open surfaces. We conduct this experiment on the \textit{dress} category from the \DFD~\cite{zhu2020deep}. We randomly select 116 objects as the training set and 25 objects as the test set.
All experiments are compared with the SOTA methods for better verification.
{To avoid thin closed reconstructions during the training process, we employ a smaller learning rate for the SDF-Net and a larger learning rate for the Validity-Net.}
Please refer to the implementation of \netName{} in the supplementary.
\vspace{-3mm}

\begin{figure*}[htb]
	\vspace{0.0mm}
	\begin{minipage}[t]{0.13\textwidth}
		\centering
		\includegraphics[width=0.9\textwidth]{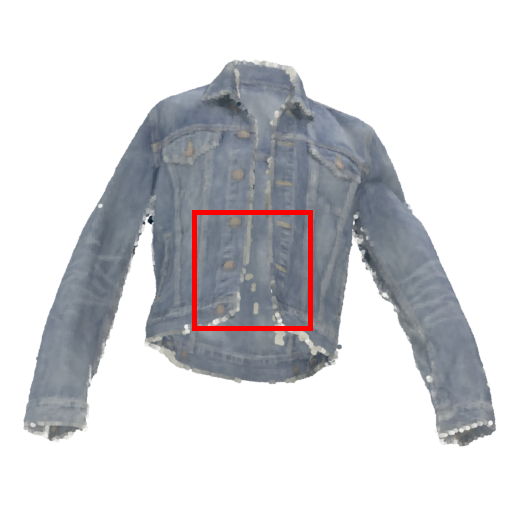}
	\end{minipage}
	\begin{minipage}[t]{0.06\textwidth}
		\centering
		\includegraphics[width=1.0\textwidth]{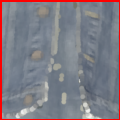}
	\end{minipage}
	\begin{minipage}[t]{0.13\textwidth}
		\centering
		\includegraphics[width=0.9\textwidth]{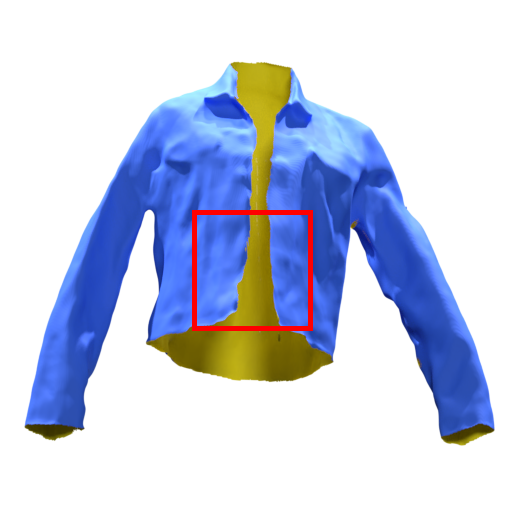}
	\end{minipage}
	\begin{minipage}[t]{0.06\textwidth}
		\centering
		\includegraphics[width=1.0\textwidth]{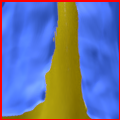}
	\end{minipage}
	\begin{minipage}[t]{0.13\textwidth}
		\centering
		\includegraphics[width=0.9\textwidth]{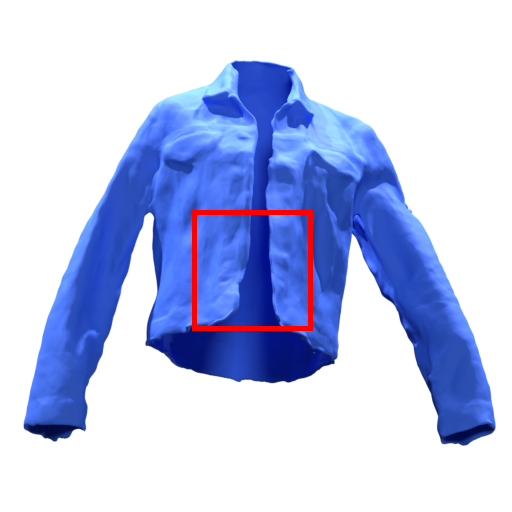}
	\end{minipage}
	\begin{minipage}[t]{0.06\textwidth}
		\centering
		\includegraphics[width=1.0\textwidth]{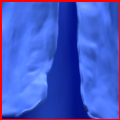}
	\end{minipage}
	\begin{minipage}[t]{0.13\textwidth}
		\centering
		\includegraphics[width=0.9\textwidth]{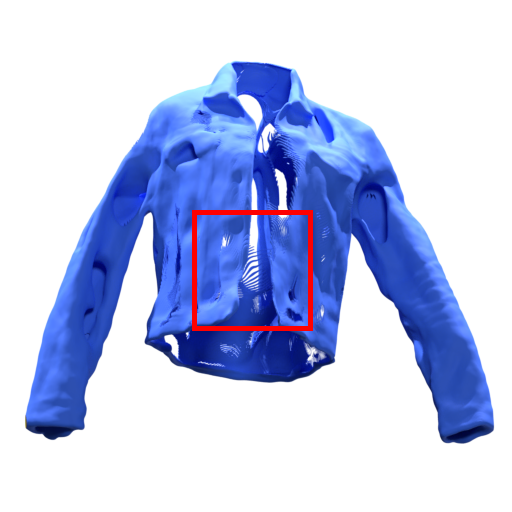}
	\end{minipage}
	\begin{minipage}[t]{0.06\textwidth}
		\centering
		\includegraphics[width=1.0\textwidth]{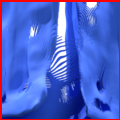}
	\end{minipage}
	\begin{minipage}[t]{0.13\textwidth}
		\centering
		\includegraphics[width=0.9\textwidth]{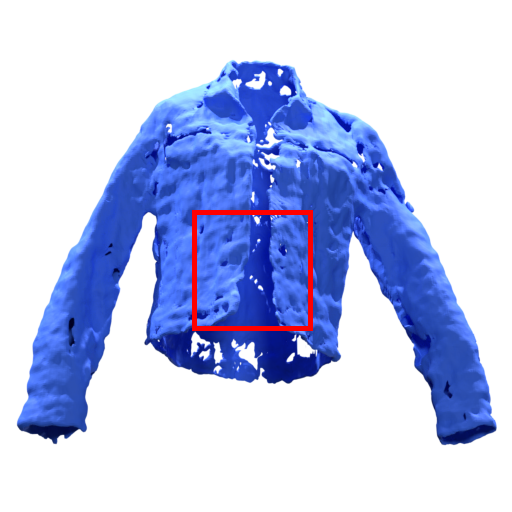}
	\end{minipage}
	\begin{minipage}[t]{0.06\textwidth}
		\centering
		\includegraphics[width=1.0\textwidth]{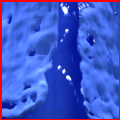}
	\end{minipage}
	\\
	\vspace{0.0mm}
	\begin{minipage}[t]{0.13\textwidth}
		\centering
		\includegraphics[width=0.9\textwidth]{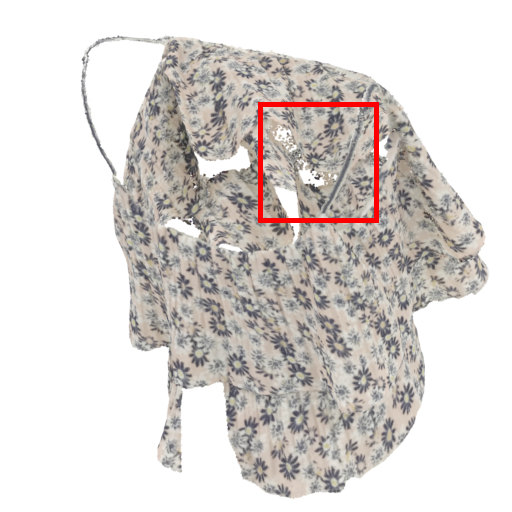}
	\end{minipage}
	\begin{minipage}[t]{0.06\textwidth}
		\centering
		\includegraphics[width=1.0\textwidth]{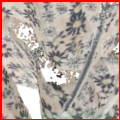}
	\end{minipage}
	\begin{minipage}[t]{0.13\textwidth}
		\centering
		\includegraphics[width=0.9\textwidth]{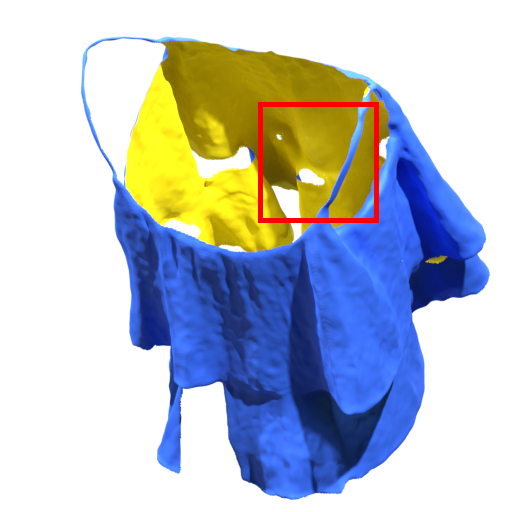}
	\end{minipage}
	\begin{minipage}[t]{0.06\textwidth}
		\centering
		\includegraphics[width=1.0\textwidth]{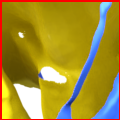}
	\end{minipage}
	\begin{minipage}[t]{0.13\textwidth}
		\centering
		\includegraphics[width=0.9\textwidth]{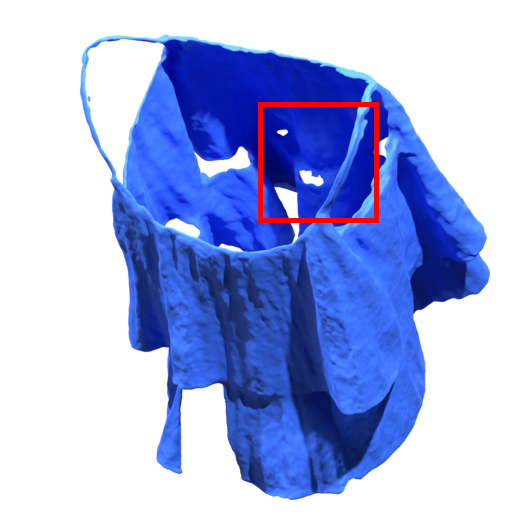}
	\end{minipage}
	\begin{minipage}[t]{0.06\textwidth}
		\centering
		\includegraphics[width=1.0\textwidth]{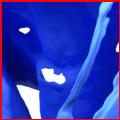}
	\end{minipage}
	\begin{minipage}[t]{0.13\textwidth}
		\centering
		\includegraphics[width=0.9\textwidth]{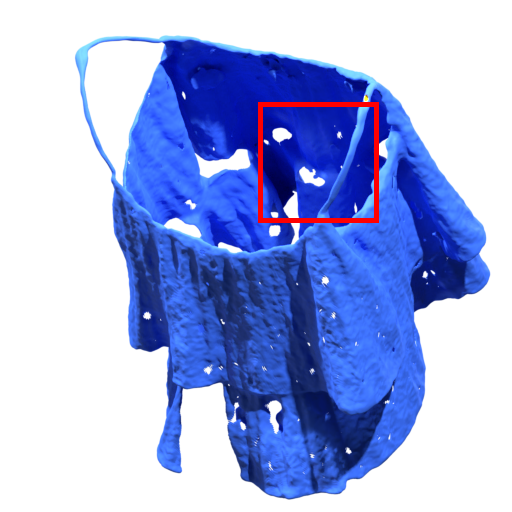}
	\end{minipage}
	\begin{minipage}[t]{0.06\textwidth}
		\centering
		\includegraphics[width=1.0\textwidth]{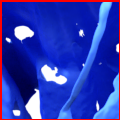}
	\end{minipage}
	\begin{minipage}[t]{0.13\textwidth}
		\centering
		\includegraphics[width=0.9\textwidth]{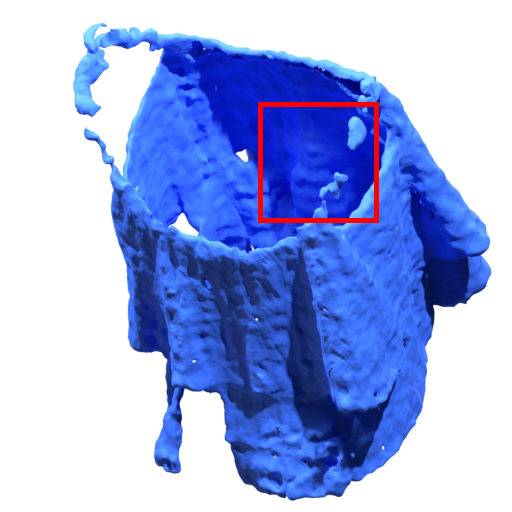}
	\end{minipage}
	\begin{minipage}[t]{0.06\textwidth}
		\centering
		\includegraphics[width=1.0\textwidth]{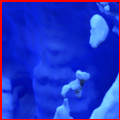}
	\end{minipage}
	\\
	\vspace{0.0mm}
	\begin{minipage}[t]{0.13\textwidth}
		\centering
		\includegraphics[width=0.9\textwidth]{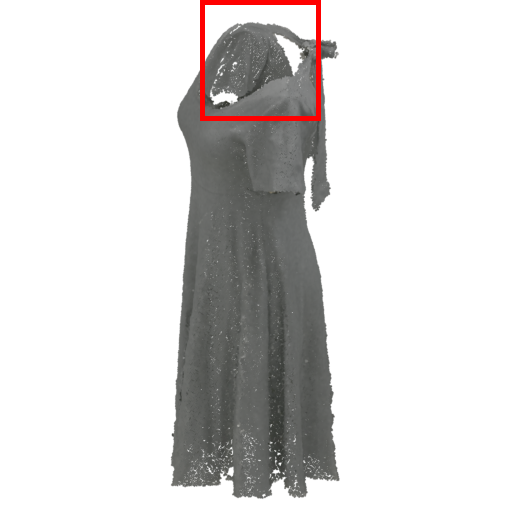}
	\end{minipage}
	\begin{minipage}[t]{0.06\textwidth}
		\centering
		\includegraphics[width=1.0\textwidth]{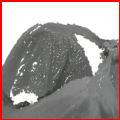}
	\end{minipage}
	\begin{minipage}[t]{0.13\textwidth}
		\centering
		\includegraphics[width=0.9\textwidth]{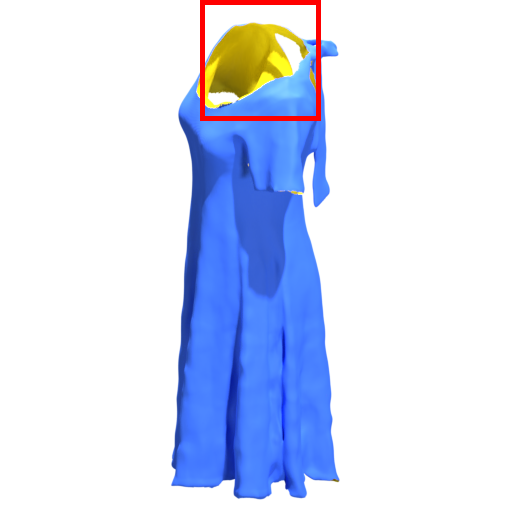}
	\end{minipage}
	\begin{minipage}[t]{0.06\textwidth}
		\centering
		\includegraphics[width=1.0\textwidth]{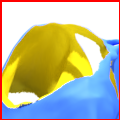}
	\end{minipage}
	\begin{minipage}[t]{0.13\textwidth}
		\centering
		\includegraphics[width=0.9\textwidth]{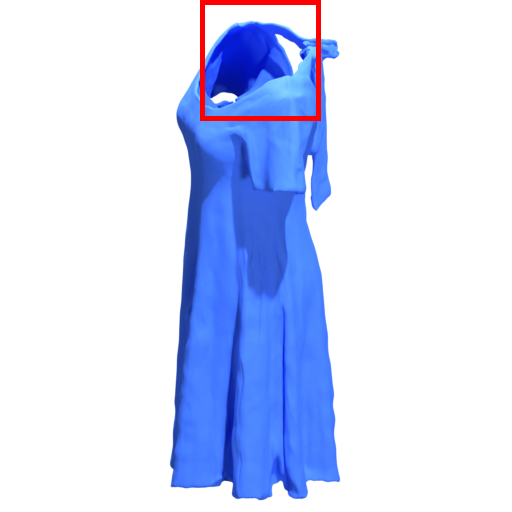}
	\end{minipage}
	\begin{minipage}[t]{0.06\textwidth}
		\centering
		\includegraphics[width=1.0\textwidth]{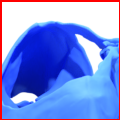}
	\end{minipage}
	\begin{minipage}[t]{0.13\textwidth}
		\centering
		\includegraphics[width=0.9\textwidth]{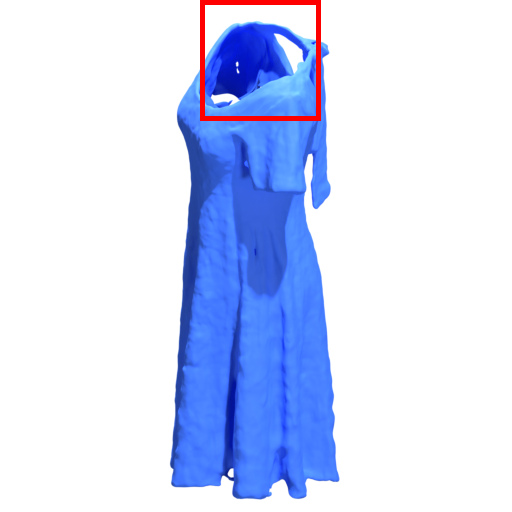}
	\end{minipage}
	\begin{minipage}[t]{0.06\textwidth}
		\centering
		\includegraphics[width=1.0\textwidth]{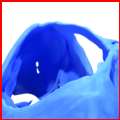}
	\end{minipage}
	\begin{minipage}[t]{0.13\textwidth}
		\centering
		\includegraphics[width=0.9\textwidth]{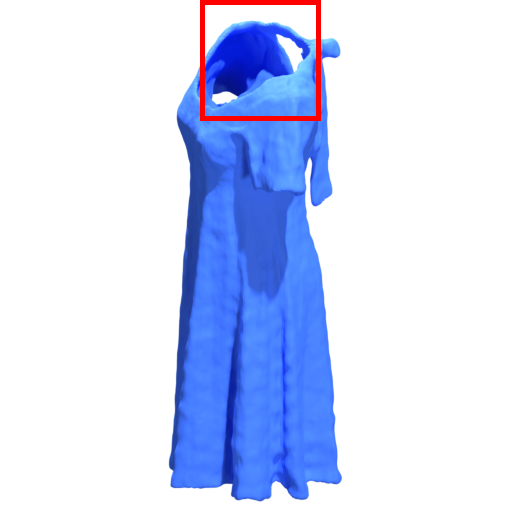}
	\end{minipage}
	\begin{minipage}[t]{0.06\textwidth}
		\centering
		\includegraphics[width=1.0\textwidth]{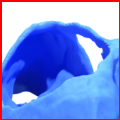}
	\end{minipage}
	\\
	\vspace{0.0mm}
	\begin{minipage}[t]{0.13\textwidth}
		\centering
		\includegraphics[width=0.9\textwidth]{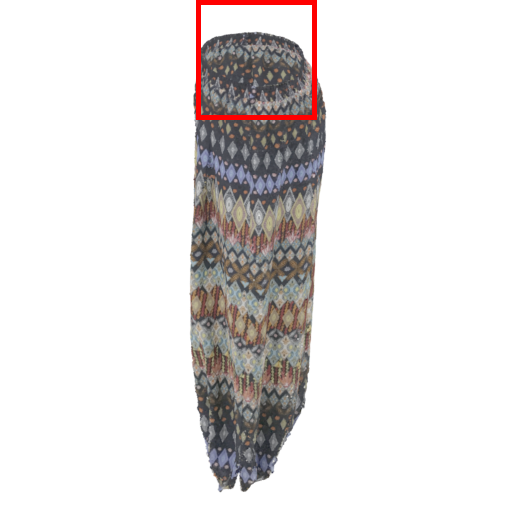}
	\end{minipage}
	\begin{minipage}[t]{0.06\textwidth}
		\centering
		\includegraphics[width=1.0\textwidth]{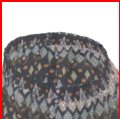}
	\end{minipage}
	\begin{minipage}[t]{0.13\textwidth}
		\centering
		\includegraphics[width=0.9\textwidth]{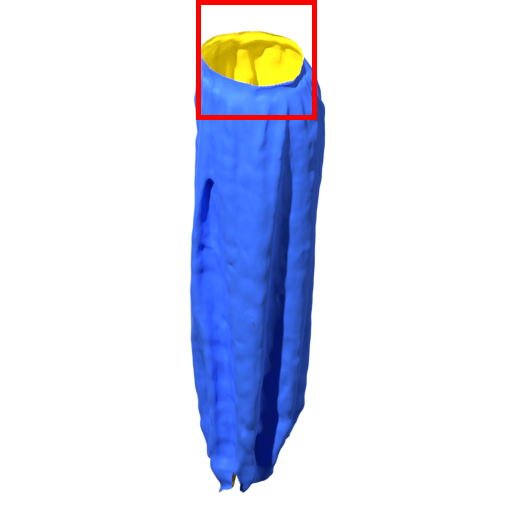}
	\end{minipage}
	\begin{minipage}[t]{0.06\textwidth}
		\centering
		\includegraphics[width=1.0\textwidth]{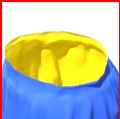}
	\end{minipage}
	\begin{minipage}[t]{0.13\textwidth}
		\centering
		\includegraphics[width=0.9\textwidth]{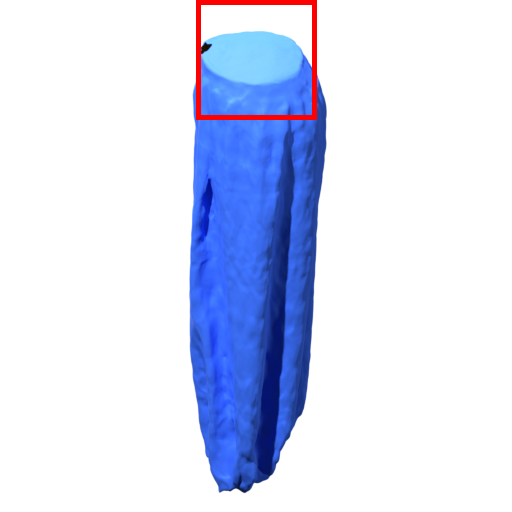}
	\end{minipage}
	\begin{minipage}[t]{0.06\textwidth}
		\centering
		\includegraphics[width=1.0\textwidth]{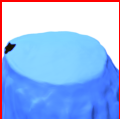}
	\end{minipage}
	\begin{minipage}[t]{0.13\textwidth}
		\centering
		\includegraphics[width=0.9\textwidth]{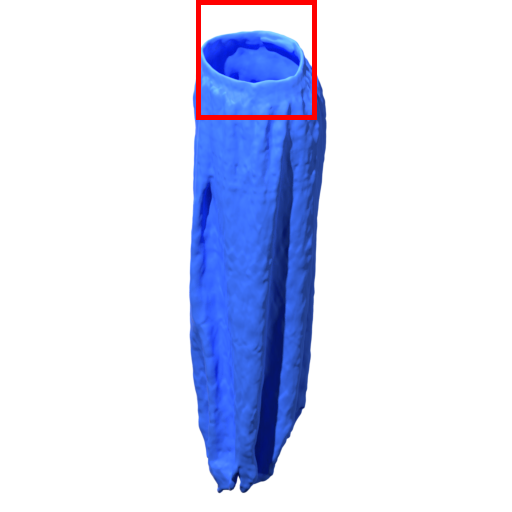}
	\end{minipage}
	\begin{minipage}[t]{0.06\textwidth}
		\centering
		\includegraphics[width=1.0\textwidth]{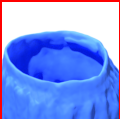}
	\end{minipage}
	\begin{minipage}[t]{0.13\textwidth}
		\centering
		\includegraphics[width=0.9\textwidth]{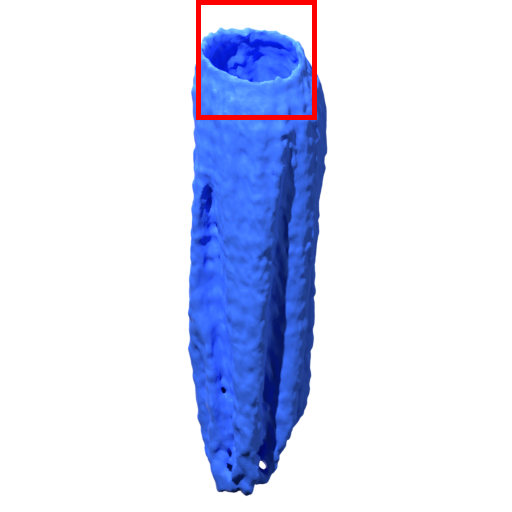}
	\end{minipage}
	\begin{minipage}[t]{0.06\textwidth}
		\centering
		\includegraphics[width=1.0\textwidth]{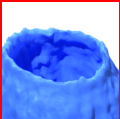}
	\end{minipage}
	\\
	\vspace{0.0mm}
	\begin{minipage}[t]{0.13\textwidth}
		\centering
		\subfloat[GT]{\includegraphics[width=0.9\textwidth]{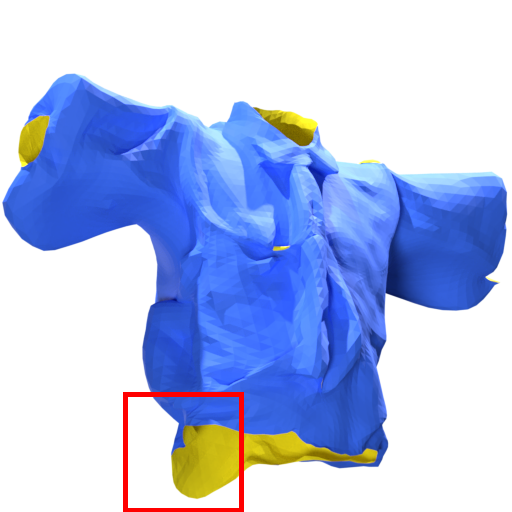}}
	\end{minipage}
	\begin{minipage}[t]{0.06\textwidth}
		\centering
		\includegraphics[width=1.0\textwidth]{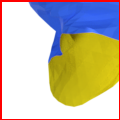}
	\end{minipage}
	\begin{minipage}[t]{0.13\textwidth}
		\centering
		\subfloat[Ours]{\includegraphics[width=0.9\textwidth]{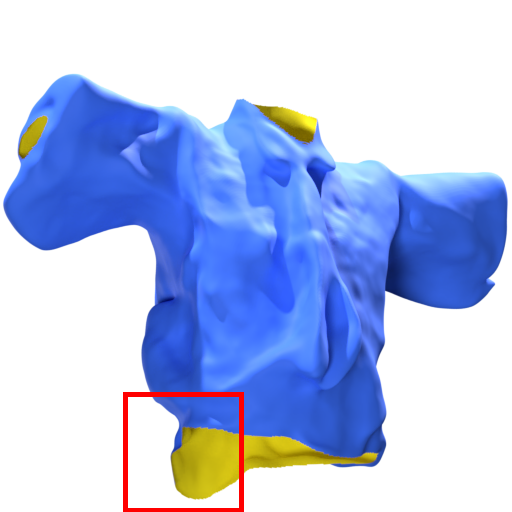}}
	\end{minipage}
	\begin{minipage}[t]{0.06\textwidth}
		\centering
		\includegraphics[width=1.0\textwidth]{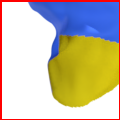}
	\end{minipage}
	\begin{minipage}[t]{0.13\textwidth}
		\centering
		\subfloat[NeuS]{\includegraphics[width=0.9\textwidth]{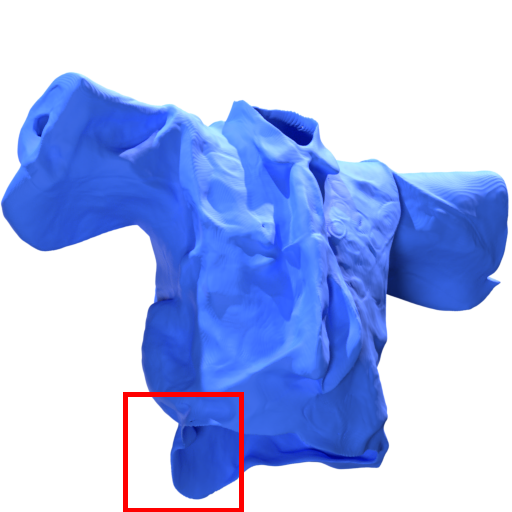}}
	\end{minipage}
	\begin{minipage}[t]{0.06\textwidth}
		\centering
		\includegraphics[width=1.0\textwidth]{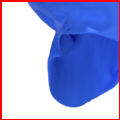}
	\end{minipage}
	\begin{minipage}[t]{0.13\textwidth}
		\centering
		\subfloat[IDR]{\includegraphics[width=0.9\textwidth]{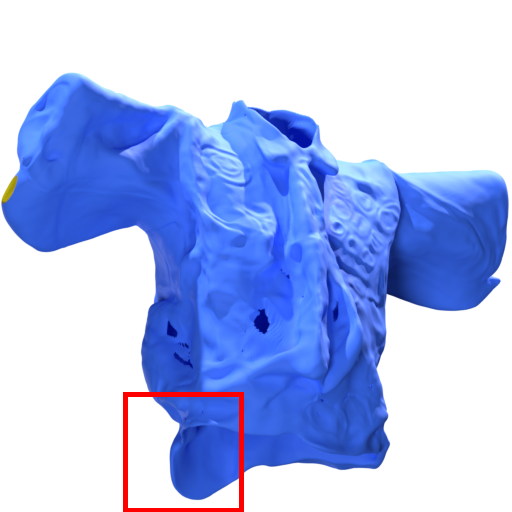}}
	\end{minipage}
	\begin{minipage}[t]{0.06\textwidth}
		\centering
		\includegraphics[width=1.0\textwidth]{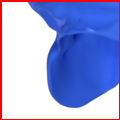}
	\end{minipage}
	\begin{minipage}[t]{0.13\textwidth}
		\centering
		\subfloat[HFS]{\includegraphics[width=0.9\textwidth]{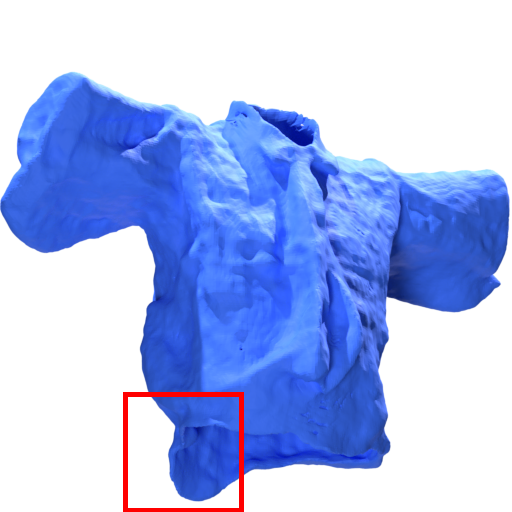}}
	\end{minipage}
	\begin{minipage}[t]{0.06\textwidth}
		\centering
		\includegraphics[width=1.0\textwidth]{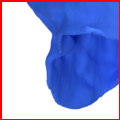}
	\end{minipage}
\vspace{-1em}
\caption{Comparisons on open surface reconstruction. Row 1 -- 4 are evaluated on \DFD~\cite{zhu2020deep} and Row 5 is evaluated on \MGN~\cite{bhatnagar2019mgn}. \modelName~is able to reconstruct high-fidelity open surfaces while NeuS~\cite{wang2021neus}, IDR~\cite{yariv2020idr} and HFS~\cite{wang2022hfneus} fail to recover the correct topologies.}
\vspace{-1em}
\label{fig:comparison_open}
\end{figure*}

\vspace{-0.5em}


\vspace{-0.5mm}

\paragraph{Implementation details.}
For the reconstruction experiments on open surfaces, we render the  ground truth point clouds from \DFD~\cite{zhu2020deep} with Pytorch3D~\cite{ravi2020pytorch3d} at a resolution of $256^2$. To get diverse supervision data, we uniformly sample 648 and 64 viewpoints on the unit sphere for \DFD~and \MGN~(MGN), respectively.
For the single view reconstruction experiment, we uniformly sample 64 viewpoints on the unit sphere as the camera positions.
We use an ResNet-18~\cite{He_2016_CVPR} encoder to predict a latent code $\textbf{z}$ describing the surface's geometry and color. We then use the concatenation of $\{\textbf{z}, \pt\}$ as the input to \netName~(decoder) to evaluate the SDF, validity, and color at the query positions. We optimize the autoencoder by comparing the 2D rendering and the ground truth image. 
In the evaluation stage, we accept a single image as the input and directly export the evaluated SDF and validity as 3D mesh.

\vspace{-1em}
\paragraph{Evaluations.} For multiview reconstruction on watertight surfaces, we measure the Chamfer Distance (CD) with \textit{DTU MVS 2014 evaluation toolkit}~\cite{dtu}. For the reconstruction experiments on open surfaces, we measure the CD with the \textit{PCU} Library~\cite{point-cloud-utils}. For all the experiments, we evaluate the result meshes at resolution $512^3$. 
\vspace{-1.5mm}
\subsection{Multiview Reconstruction on Closed Surfaces}
\vspace{-0.5em}


We compare our approach with the state-of-the-art volume and surface rendering based methods - HFS~\cite{wang2022hfneus}, NeuS~\cite{wang2021neus} and IDR~\cite{yariv2020idr}, and a classic mesh reconstruction and novel view synthesis method -- NeRF~\cite{mildenhall2020nerf}. We report the quantitative results in Table~\ref{table:comparison_watertight}.

We also show visual comparison with a widely-used MVS method: COLMAP~\cite{schoenberger2016sfm, schoenberger2016mvs}. 
We show qualitative results in Fig.~\ref{fig:comparison_watertight}. The results reconstructed with the proposed method show comparable quality compared with the state-of-the-art.



\begin{table}[h]
    \small
    \centering
    \begin{tabular}{c|c|c|c|c|c}
        \hline
        CD$\downarrow$ & Ours & NeuS & IDR & NeRF & HFS\\
        \hline
        \hline
        scan 55 & $0.47$ & $0.38$ & $0.48$ & $0.66$ & $\textbf{0.37}$\\
        scan 69 & $0.84$ & $\textbf{0.60}$ & $0.77$ & $1.50$ & $0.66$\\
        scan 83 & $1.28$ & $1.43$ & $1.33$ & $\textbf{1.20}$ & $1.27$\\
        scan 97 & $1.09$ & $\textbf{0.96}$ & $1.16$ & $1.96$ & $1.00$\\
        scan 105 & $\textbf{0.75}$ & $0.78$ & $0.76$ & $1.27$ & $0.86$\\
        scan 106 & $0.76$ & $\textbf{0.52}$ & $0.67$ & $0.66$ & $0.57$\\
        scan 110 & $\textbf{0.80}$ & $1.44$ & $0.90$ & $2.61$ & $1.24$\\
        scan 114 & $0.38$ & $\textbf{0.36}$ & $0.42$ & $1.04$ & $0.41$\\
        scan 118 & $0.56$ & $\textbf{0.46}$ & $0.51$ & $1.13$ & $0.52$\\
        scan 122 & $0.55$ & $\textbf{0.49}$ & $0.53$ & $0.99$ & $\textbf{0.49}$\\
        \hline
        \hline
        average & $0.749$ & $0.742$ & $0.753$ & $1.302$ & $\textbf{0.741}$\\
        \hline
    \end{tabular}
    \caption{Quantitative evals on real-world object reconstruction.
    }
    \label{table:comparison_watertight}
\end{table}

\begin{table}[htbp]
    \small
    \centering
    \begin{tabular}{c|c|c|c|c|c}
        \hline
        & CD ($\times 10^{-3}$) $\downarrow$ & Ours & NeuS & IDR & HFS \\
        \hline
        \hline
        \multirow{9}{*}{D3D} 
        &long slv upper & \textbf{4.483} & $6.864$ & $11.494$ & $9.695$\\
        &short slv upper & \textbf{4.517} & $6.048 $ & $9.043$ & $7.800$\\
        &no slv upper & \textbf{3.418} & $4.856 $ & $17.710$ & $8.576$\\
        &long slv dress & \textbf{4.843} & $6.135$ & $9.203$ & $8.235$\\
        &short slv dress & \textbf{4.276} & $7.951$ & $8.506$ & $7.705$\\
        &no slv dress & \textbf{3.706} & $5.406$ & $6.785$ & $7.565$\\
        &pants & \textbf{5.391} & $11.847 $ & $10.880$ & $16.205$\\
        &dress & \textbf{3.889} & $5.673 $ & $6.983$ & $11.644$\\
        \cline{2-6}
        &average & \textbf{4.315} & $6.847$ & $10.075$ & $9.678$\\
        \hline
        \hline
        \multirow{6}{*}{MGN}
        &LongCoat & \textbf{7.601} & $8.038$ & $12.058$ & $10.398$\\
        &TShirtNoCoat & \textbf{8.481} & $9.910$ & $15.709$ & $13.128$\\
        &ShirtNoCoat & \textbf{5.281} & $8.084$ & $9.509$ & $11.299$\\
        &ShortPants & \textbf{15.324} & $15.480$ & $16.329$ & $18.332$\\
        &Pants & \textbf{9.191} & $12.188$ & $19.931$ & $19.414$\\
        \cline{2-6}
        &average & \textbf{9.176} & $10.740$ & $14.707$ & $14.514$\\
        \hline
    \end{tabular}
    \vspace{-0.2em}
    \caption{Quantitative evaluation on \textit{Deep Fashion 3D
    Dataset}~(D3D)~\cite{zhu2020deep}~with chamfer distance averaged over five examples per category, and \MGN~(MGN)~\cite{bhatnagar2019mgn} with chamfer distance averaged on two examples per category.}
    \vspace{-1em}
    \label{table:comparison_open_d3d}
\end{table}
\vspace{-0.5em}
\subsection{Multiview Reconstruction on Open Surfaces}
\vspace{-0.5em}

We conduct this experiment on eight categories from Deep Fashion 3D~\cite{zhu2020deep} and five categories from the MGN dataset~\cite{bhatnagar2019mgn}. {We compare our approach with two state-of-the-art volume rendering based methods -- NeuS~\cite{wang2021neus} and HFS~\cite{wang2022hfneus}, and a surface rendering based method -- IDR~\cite{yariv2020idr}.}

We report the Chamfer Distance averaged on five examples for each category from \DFD~\cite{zhu2020deep} and report the Chamfer Distance averaged on two examples for each category from \MGN~in Table~\ref{table:comparison_open_d3d}. \modelName~generally provides lower numerical errors compared with the state-of-the-arts.
We show qualitative results in Fig.~\ref{fig:comparison_open}. {\modelName~also provides lower numerical errors in F-score. Please refer to the supplemental for the comparisons.}

In most cases, NeuS~\cite{wang2021neus} and IDR~\cite{yariv2020idr} are able to reconstruct the geometry with thick, watertight surfaces. While, for the pants in Figure~\ref{fig:comparison_open}, NeuS fails to recover the shape of the waist. \modelName{} is able to reconstruct high-fidelity open surfaces with consistent normals, including the thin straps of the camisoles and dresses.
\begin{figure}[htbp]
\centering
    \includegraphics[width=0.95\linewidth]{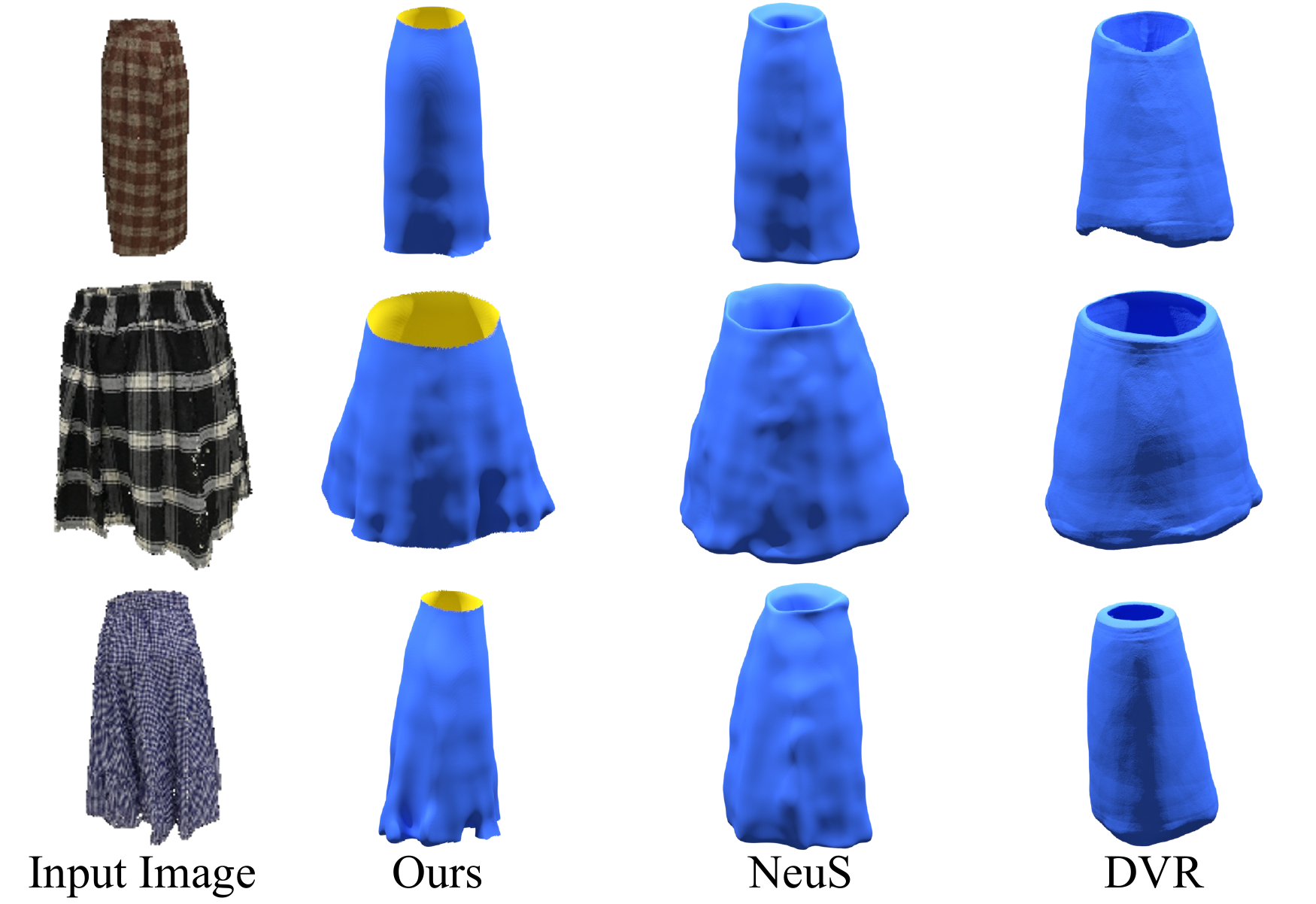}
\vspace{-0.5em}
\caption{With given single-view images, ours predicts
accurate 3D geometry of arbitrary shapes with the autoencoder. \modelName{} achieves CD = $0.0771$ averaged on the 25
objects from the test set, which outperforms NeuS~\cite{wang2021neus} (CD =
$0.0778$) and DVR~\cite{dvr} (CD = $0.0789$). }
\vspace{-1.5em}
\label{fig:comparison_singleview_reconstruction}
\end{figure}

\vspace{-0.5em}
\subsection{Single View Reconstruction on Open Surfaces}
\vspace{-0.5em}
We construct an autoencoder, which accepts a single image as the input, and exports the 3D mesh as the output. For this experiment, we compare our approach against the state-of-the-art single-view reconstruction method: DVR~\cite{dvr} and the volume rendering based method: NeuS~\cite{wang2021neus}.


The qualitative results is shown in Fig~\ref{fig:comparison_singleview_reconstruction}.
Our method is able to infer accurate 3D shape representations from single-view images when only using 2D multi-view images and object masks as supervision.
Qualitatively, in contrast to the DVR~\cite{dvr} and NeuS~\cite{wang2021neus} autoencoder, our method is able to reconstruct open surfaces.
Quantitatively, our method achieves CD = $0.0771$, which outperforms NeuS (CD = $0.0778$) and DVR (CD = $0.0789$) averaged on all the 25 objects from the test set.

\subsection{Ablation Studies}
\vspace{-0.5em}
\begin{figure}[htb]
    \centering
        \includegraphics[width=1.0\linewidth]{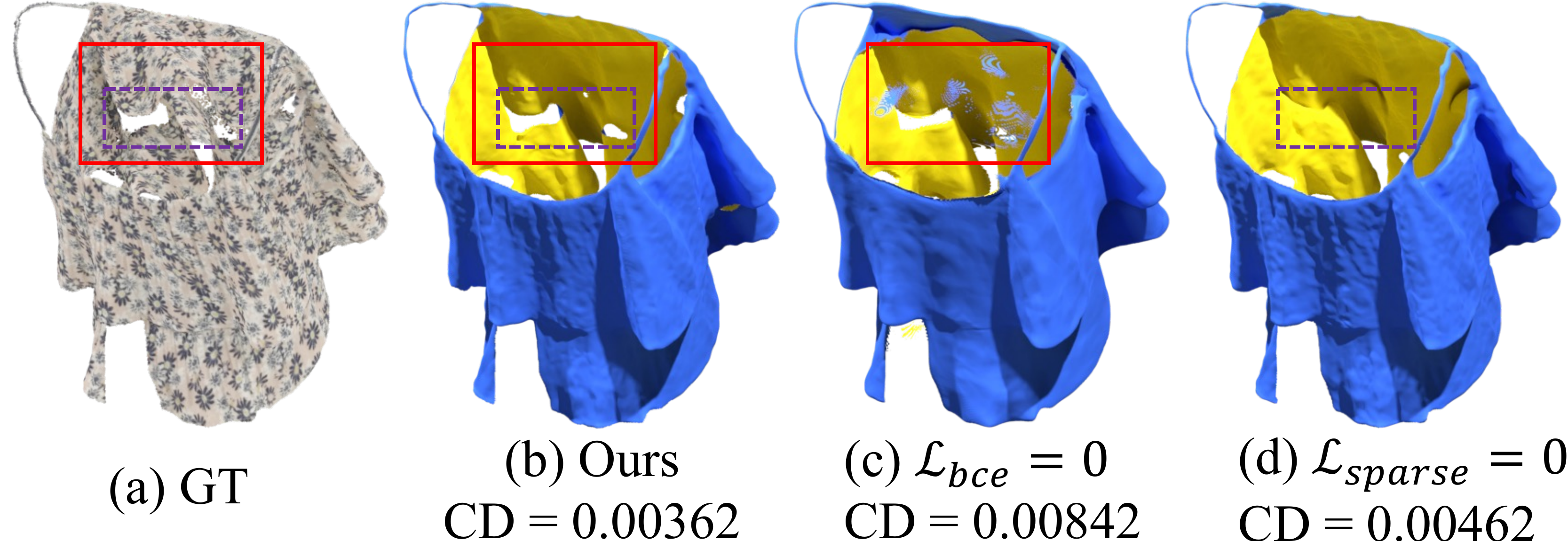}
    \vspace{-1.5em}
    \caption{Ablation study on the regularizations about validity.
    }
\vspace{-2em}
\label{fig:ablation_validity}
\end{figure}

\paragraph{Regularizations on validity.}
We conduct an ablation study on the regularizations about validity, i.e. $\mathcal{L}_{bce}$ and $\mathcal{L}_{sparse}$.
As shown in Figure~\ref{fig:ablation_validity} (c), by setting $\mathcal{L}_{bce}=0$, the renderer tends to generate rendering probability between 0 and 1, thus resulting in noisy faces in the output mesh; as shown in Figure~\ref{fig:ablation_validity} (d), by setting $\mathcal{L}_{sparse}=0$, the renderer will keep the redundant surfaces, instead of learning a validity space as sparse as possible.


\begin{figure}[htb]
    \begin{minipage}[t]{.09\textwidth}
        \centering
        \subfloat[GT]{\includegraphics[width=\textwidth]{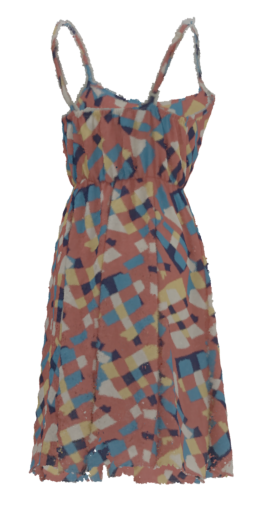}}
    \end{minipage}
    \begin{minipage}[t]{.09\textwidth}
        \centering
        \subfloat[64 views\\CD = 0.00568]{\includegraphics[width=\textwidth]{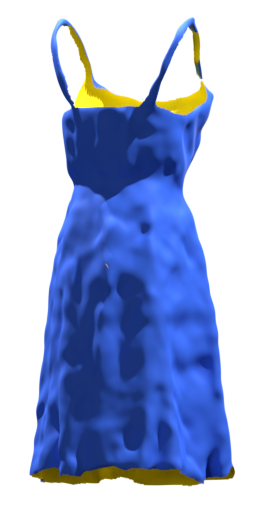}}
    \end{minipage}
    \begin{minipage}[t]{.09\textwidth}
        \centering
        \subfloat[32 views\\CD = 0.00632]{\includegraphics[width=\textwidth]{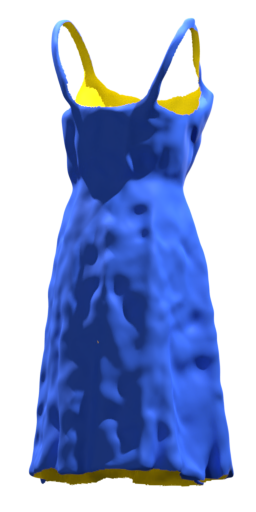}}
    \end{minipage}
    \begin{minipage}[t]{.09\textwidth}
        \centering
        \subfloat[16 views\\CD = 0.00763]{\includegraphics[width=\textwidth]{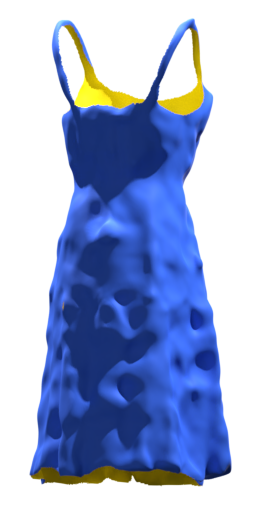}}
    \end{minipage}
    \begin{minipage}[t]{.09\textwidth}
        \centering
        \subfloat[8 views\\CD = 0.01326]{\includegraphics[width=\textwidth]{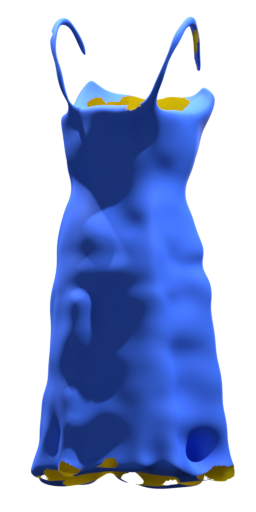}}
    \end{minipage}
    \vspace{-1em}
    \caption{Ablation study on multi-view reconstruction with different number of views.}
\vspace{-1em}
\label{fig:ablation_n_views}
\end{figure}

\paragraph{Reconstruct with different number of views.}
We additionally show results on reconstruction with different number of views. As shown in Figure~\ref{fig:ablation_n_views}, our method is able to reconstruct open surfaces even with sparse viewpoints. The reconstruction quality improves with the increase of views, quantitively and qualitatively.





\section{Discussions and Conclusions}
\vspace{-1em}
\begin{figure}[htbp]
\centering
    \includegraphics[width=0.95\linewidth]{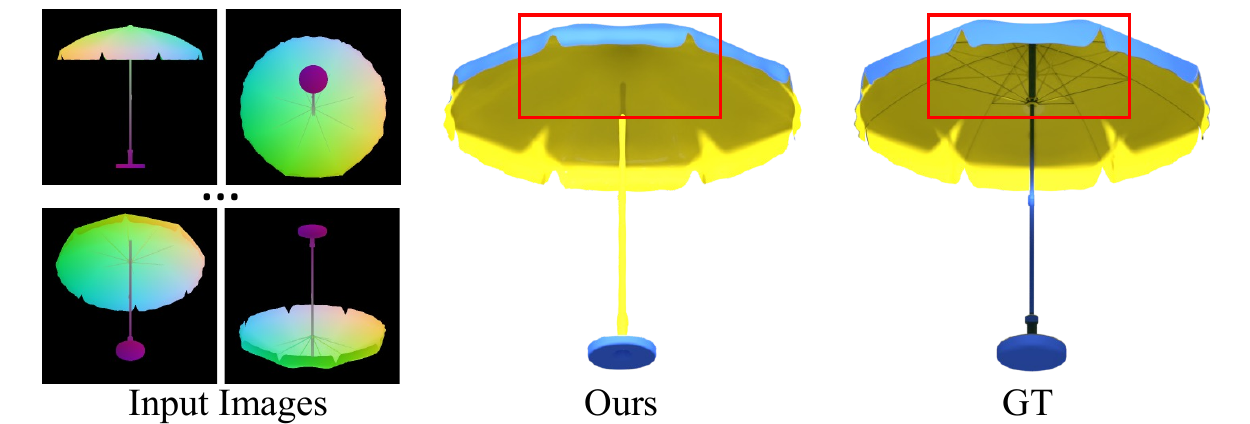}
\vspace{-0.5em}
\caption{A failure case: our method fails to reconstruct the thin stretchers of the umbrella.}
\vspace{-2em}
\label{fig:failure_case}
\end{figure}

\paragraph{Limitations and Future works.}
{
Figure~\ref{fig:comparison_open} Row 1 illustrates that the void space around the collar is obscured by limited input views and occlusions, causing it to be mistakenly connected with the jacket's main body by all reconstruction methods.
In addition, the output normal orientation of \modelName{} is influenced by the initialization of SDF-Net. 
However, by using geometric initialization~\cite{igr}, we can easily set the surface to have an initial outward normal distribution, allowing us to obtain out-facing 3D reconstructions. 
}
Finally, our method has difficulty in reconstructing very thin closed surfaces, such as the umbrella stretchers in Figure~\ref{fig:failure_case}.

A future avenue would be introducing more advanced adaptive sampling and weighting mechanisms to reconstruct highly intricate structures. 
Another direction for future work is extending \modelName{} to handle in-the-wild images without camera parameters, which can enable our method to leverage more image sources for 3D unsupervised learning.
\vspace{-0.5em}

\paragraph{Conclusions.} We have proposed \modelName, a novel approach to reconstruct high-fidelity arbitrary surfaces with consistent normals from multi-view images.
By representing the surface as a combination of the SDF and the validity probability, we develop a new volume rendering method for learning the implicit representation. Our method outperforms the state-of-the-art neural surface reconstruction methods on reconstructing open surfaces and achieves comparative results on reconstructing watertight surfaces.

{\small
\bibliographystyle{ieee_fullname}
\bibliography{egbib}

\begin{thebibliography}{10}\itemsep=-1pt

\bibitem{achlioptas2018learning}
Panos Achlioptas, Olga Diamanti, Ioannis Mitliagkas, and Leonidas Guibas.
\newblock Learning representations and generative models for 3d point clouds,
  2018.

\bibitem{Atzmon_2020_CVPR}
Matan Atzmon and Yaron Lipman.
\newblock Sal: Sign agnostic learning of shapes from raw data.
\newblock In {\em IEEE/CVF Conference on Computer Vision and Pattern
  Recognition (CVPR)}, June 2020.

\bibitem{BPA}
F. Bernardini, J. Mittleman, H. Rushmeier, C. Silva, and G. Taubin.
\newblock The ball-pivoting algorithm for surface reconstruction.
\newblock {\em IEEE Transactions on Visualization and Computer Graphics},
  5(4):349--359, 1999.

\bibitem{bhatnagar2019mgn}
Bharat~Lal Bhatnagar, Garvita Tiwari, Christian Theobalt, and Gerard Pons-Moll.
\newblock Multi-garment net: Learning to dress 3d people from images.
\newblock In {\em {IEEE} International Conference on Computer Vision ({ICCV})}.
  {IEEE}, oct 2019.

\bibitem{shapenet2015}
Angel~X. Chang, Thomas Funkhouser, Leonidas Guibas, Pat Hanrahan, Qixing Huang,
  Zimo Li, Silvio Savarese, Manolis Savva, Shuran Song, Hao Su, Jianxiong Xiao,
  Li Yi, and Fisher Yu.
\newblock {ShapeNet: An Information-Rich 3D Model Repository}.
\newblock Technical Report arXiv:1512.03012 [cs.GR], Stanford University ---
  Princeton University --- Toyota Technological Institute at Chicago, 2015.

\bibitem{chen_2022_3psdf}
Weikai Chen, Cheng Lin, Weiyang Li, and Bo Yang.
\newblock 3psdf: Three-pole signed distance function for learning surfaces with
  arbitrary topologies.
\newblock {\em Proceedings of the IEEE/CVF Conference on Computer Vision and
  Pattern Recognition}, June 2022.

\bibitem{NEURIPS2019_f5ac21cd}
Wenzheng Chen, Huan Ling, Jun Gao, Edward Smith, Jaakko Lehtinen, Alec
  Jacobson, and Sanja Fidler.
\newblock Learning to predict 3d objects with an interpolation-based
  differentiable renderer.
\newblock In H. Wallach, H. Larochelle, A. Beygelzimer, F. d\textquotesingle
  Alch\'{e}-Buc, E. Fox, and R. Garnett, editors, {\em Advances in Neural
  Information Processing Systems}, volume~32. Curran Associates, Inc., 2019.

\bibitem{chen2022ndc}
Zhiqin Chen, Andrea Tagliasacchi, Thomas Funkhouser, and Hao Zhang.
\newblock Neural dual contouring.
\newblock {\em ACM Transactions on Graphics (Special Issue of SIGGRAPH)},
  41(4), 2022.

\bibitem{sdf0}
Zhiqin Chen and Hao Zhang.
\newblock Learning implicit fields for generative shape modeling.
\newblock In {\em Proceedings of the IEEE/CVF Conference on Computer Vision and
  Pattern Recognition}, pages 5939--5948, 2019.

\bibitem{chibane2020ndf}
Julian Chibane, Aymen Mir, and Gerard Pons-Moll.
\newblock Neural unsigned distance fields for implicit function learning.
\newblock In {\em Advances in Neural Information Processing Systems
  ({NeurIPS})}, December 2020.

\bibitem{3dr2n2}
Christopher~B. Choy, Danfei Xu, JunYoung Gwak, Kevin Chen, and Silvio Savarese.
\newblock 3d-r2n2: A unified approach for single and multi-view 3d object
  reconstruction.
\newblock In Bastian Leibe, Jiri Matas, Nicu Sebe, and Max Welling, editors,
  {\em Computer Vision -- ECCV 2016}, pages 628--644, Cham, 2016. Springer
  International Publishing.

\bibitem{8099747}
Haoqiang Fan, Hao Su, and Leonidas Guibas.
\newblock A point set generation network for 3d object reconstruction from a
  single image.
\newblock In {\em 2017 IEEE Conference on Computer Vision and Pattern
  Recognition (CVPR)}, pages 2463--2471, 2017.

\bibitem{genova2018unsupervised}
Kyle Genova, Forrester Cole, Aaron Maschinot, Aaron Sarna, Daniel Vlasic, and
  William~T Freeman.
\newblock Unsupervised training for 3d morphable model regression.
\newblock In {\em Proceedings of the IEEE Conference on Computer Vision and
  Pattern Recognition}, pages 8377--8386, 2018.

\bibitem{8578972}
Kyle Genova, Forrester Cole, Aaron Maschinot, Aaron Sarna, Daniel Vlasic, and
  William~T. Freeman.
\newblock Unsupervised training for 3d morphable model regression.
\newblock In {\em 2018 IEEE/CVF Conference on Computer Vision and Pattern
  Recognition}, pages 8377--8386, 2018.

\bibitem{igr}
Amos Gropp, Lior Yariv, Niv Haim, Matan Atzmon, and Yaron Lipman.
\newblock Implicit geometric regularization for learning shapes.
\newblock In {\em Proceedings of Machine Learning and Systems 2020}, pages
  3569--3579. 2020.

\bibitem{atlasnet}
Thibault Groueix, Matthew Fisher, Vladimir~G. Kim, Bryan Russell, and Mathieu
  Aubry.
\newblock {AtlasNet: A Papier-M\^ach\'e Approach to Learning 3D Surface
  Generation}.
\newblock In {\em Proceedings IEEE Conf. on Computer Vision and Pattern
  Recognition (CVPR)}, 2018.

\bibitem{guillard2021meshudf}
Benoit Guillard, Federico Stella, and Pascal Fua.
\newblock Meshudf: Fast and differentiable meshing of unsigned distance field
  networks.
\newblock In {\em European Conference on Computer Vision}, 2022.

\bibitem{He_2016_CVPR}
Kaiming He, Xiangyu Zhang, Shaoqing Ren, and Jian Sun.
\newblock Deep residual learning for image recognition.
\newblock In {\em Proceedings of the IEEE Conference on Computer Vision and
  Pattern Recognition (CVPR)}, June 2016.

\bibitem{zhu2020deep}
Zhu Heming, Cao Yu, Jin Hang, Chen Weikai, Du Dong, Wang Zhangye, Cui Shuguang,
  and Han Xiaoguang.
\newblock Deep fashion3d: A dataset and benchmark for 3d garment reconstruction
  from single images.
\newblock In {\em Computer Vision -- ECCV 2020}, pages 512--530. Springer
  International Publishing, 2020.

\bibitem{dtu}
Rasmus Jensen, Anders Dahl, George Vogiatzis, Engil Tola, and Henrik Aanæs.
\newblock Large scale multi-view stereopsis evaluation.
\newblock In {\em 2014 IEEE Conference on Computer Vision and Pattern
  Recognition}, pages 406--413, 2014.

\bibitem{jiang2020sdfdiff}
Yue Jiang, Dantong Ji, Zhizhong Han, and Matthias Zwicker.
\newblock Sdfdiff: Differentiable rendering of signed distance fields for 3d
  shape optimization.
\newblock In {\em The IEEE/CVF Conference on Computer Vision and Pattern
  Recognition (CVPR)}, June 2020.

\bibitem{kato2018neural}
Hiroharu Kato, Yoshitaka Ushiku, and Tatsuya Harada.
\newblock Neural 3d mesh renderer.
\newblock In {\em Proceedings of the IEEE conference on computer vision and
  pattern recognition}, pages 3907--3916, 2018.

\bibitem{li2022voxsurf}
Hai Li, Xingrui Yang, Hongjia Zhai, Yuqian Liu, Hujun Bao, and Guofeng Zhang.
\newblock Vox-surf: Voxel-based implicit surface representation.
\newblock {\em IEEE Transactions on Visualization and Computer Graphics}, pages
  1--12, 2022.

\bibitem{lin2018learning}
Chen-Hsuan Lin, Chen Kong, and Simon Lucey.
\newblock Learning efficient point cloud generation for dense 3d object
  reconstruction.
\newblock In {\em AAAI Conference on Artificial Intelligence ({AAAI})}, 2018.

\bibitem{softras}
Shichen Liu, Weikai Chen, Tianye Li, and Hao Li.
\newblock Soft rasterizer: A differentiable renderer for image-based 3d
  reasoning.
\newblock In {\em 2019 IEEE/CVF International Conference on Computer Vision
  (ICCV)}, pages 7707--7716, 2019.

\bibitem{liu2019learning}
Shichen Liu, Shunsuke Saito, Weikai Chen, and Hao Li.
\newblock Learning to infer implicit surfaces without 3d supervision.
\newblock {\em Advances in Neural Information Processing Systems}, 32, 2019.

\bibitem{dist}
Shaohui Liu, Yinda Zhang, Songyou Peng, Boxin Shi, Marc Pollefeys, and Zhaopeng
  Cui.
\newblock Dist: Rendering deep implicit signed distance function with
  differentiable sphere tracing.
\newblock In {\em Proceedings of the IEEE/CVF Conference on Computer Vision and
  Pattern Recognition (CVPR)}, June 2020.

\bibitem{liu2023neudf}
Yu-Tao Liu, Li Wang, Jie Yang, Weikai Chen, Xiaoxu Meng, Bo Yang, and Lin Gao.
\newblock Neudf: Leaning neural unsigned distance fields with volume rendering.
\newblock In {\em Proceedings of the IEEE/CVF Conference on Computer Vision and
  Pattern Recognition (CVPR)}, 2023.

\bibitem{long2023neuraludf}
Xiaoxiao Long, Cheng Lin, Lingjie Liu, Yuan Liu, Peng Wang, Christian Theobalt,
  Taku Komura, and Wenping Wang.
\newblock Neuraludf: Learning unsigned distance fields for multi-view
  reconstruction of surfaces with arbitrary topologies.
\newblock In {\em Proceedings of the IEEE/CVF Conference on Computer Vision and
  Pattern Recognition (CVPR)}, 2023.

\bibitem{marching_cubes}
William~E. Lorensen and Harvey~E. Cline.
\newblock Marching cubes: A high resolution 3d surface construction algorithm.
\newblock In {\em Proceedings of the 14th Annual Conference on Computer
  Graphics and Interactive Techniques}, SIGGRAPH '87, page 163–169, New York,
  NY, USA, 1987. Association for Computing Machinery.

\bibitem{mandikal20183dlmnet}
Priyanka Mandikal, K~L Navaneet, Mayank Agarwal, and R~Venkatesh Babu.
\newblock {3D-LMNet}: Latent embedding matching for accurate and diverse 3d
  point cloud reconstruction from a single image.
\newblock In {\em Proceedings of the British Machine Vision Conference
  ({BMVC})}, 2018.

\bibitem{voxnet}
Daniel Maturana and Sebastian Scherer.
\newblock Voxnet: A 3d convolutional neural network for real-time object
  recognition.
\newblock In {\em 2015 IEEE/RSJ International Conference on Intelligent Robots
  and Systems (IROS)}, pages 922--928, 2015.

\bibitem{occnet}
Lars Mescheder, Michael Oechsle, Michael Niemeyer, Sebastian Nowozin, and
  Andreas Geiger.
\newblock Occupancy networks: Learning 3d reconstruction in function space.
\newblock In {\em Proceedings IEEE Conf. on Computer Vision and Pattern
  Recognition (CVPR)}, 2019.

\bibitem{michalkiewicz2019implicit}
Mateusz Michalkiewicz, Jhony~Kaesemodel Pontes, Dominic Jack, Mahsa
  Baktashmotlagh, and Anders Eriksson.
\newblock Implicit surface representations as layers in neural networks.
\newblock In {\em 2019 IEEE/CVF International Conference on Computer Vision
  (ICCV)}, pages 4742--4751, 2019.

\bibitem{mildenhall2020nerf}
Ben Mildenhall, Pratul~P Srinivasan, Matthew Tancik, Jonathan~T Barron, Ravi
  Ramamoorthi, and Ren Ng.
\newblock Nerf: Representing scenes as neural radiance fields for view
  synthesis.
\newblock In {\em European conference on computer vision}, pages 405--421.
  Springer, 2020.

\bibitem{niemeyer2020differentiable}
Michael Niemeyer, Lars Mescheder, Michael Oechsle, and Andreas Geiger.
\newblock Differentiable volumetric rendering: Learning implicit 3d
  representations without 3d supervision.
\newblock In {\em Proceedings of the IEEE/CVF Conference on Computer Vision and
  Pattern Recognition}, pages 3504--3515, 2020.

\bibitem{dvr}
Michael Niemeyer, Lars Mescheder, Michael Oechsle, and Andreas Geiger.
\newblock Differentiable volumetric rendering: Learning implicit 3d
  representations without 3d supervision.
\newblock In {\em Proc. IEEE Conf. on Computer Vision and Pattern Recognition
  (CVPR)}, 2020.

\bibitem{oechsle2021unisurf}
Michael Oechsle, Songyou Peng, and Andreas Geiger.
\newblock Unisurf: Unifying neural implicit surfaces and radiance fields for
  multi-view reconstruction.
\newblock In {\em Proceedings of the IEEE/CVF International Conference on
  Computer Vision}, pages 5589--5599, 2021.

\bibitem{deepsdf}
Jeong~Joon Park, Peter Florence, Julian Straub, Richard Newcombe, and Steven
  Lovegrove.
\newblock Deepsdf: Learning continuous signed distance functions for shape
  representation.
\newblock In {\em The IEEE Conference on Computer Vision and Pattern
  Recognition (CVPR)}, June 2019.

\bibitem{ravi2020pytorch3d}
Nikhila Ravi, Jeremy Reizenstein, David Novotny, Taylor Gordon, Wan-Yen Lo,
  Justin Johnson, and Georgia Gkioxari.
\newblock Accelerating 3d deep learning with pytorch3d.
\newblock {\em arXiv:2007.08501}, 2020.

\bibitem{pifuSHNMKL19}
Shunsuke Saito, , Zeng Huang, Ryota Natsume, Shigeo Morishima, Angjoo Kanazawa,
  and Hao Li.
\newblock Pifu: Pixel-aligned implicit function for high-resolution clothed
  human digitization.
\newblock {\em arXiv preprint arXiv:1905.05172}, 2019.

\bibitem{schoenberger2016sfm}
Johannes~Lutz Sch\"{o}nberger and Jan-Michael Frahm.
\newblock Structure-from-motion revisited.
\newblock In {\em Conference on Computer Vision and Pattern Recognition
  (CVPR)}, 2016.

\bibitem{schoenberger2016mvs}
Johannes~Lutz Sch\"{o}nberger, Enliang Zheng, Marc Pollefeys, and Jan-Michael
  Frahm.
\newblock Pixelwise view selection for unstructured multi-view stereo.
\newblock In {\em European Conference on Computer Vision (ECCV)}, 2016.

\bibitem{7410471}
Hang Su, Subhransu Maji, Evangelos Kalogerakis, and Erik Learned-Miller.
\newblock Multi-view convolutional neural networks for 3d shape recognition.
\newblock In {\em 2015 IEEE International Conference on Computer Vision
  (ICCV)}, pages 945--953, 2015.

\bibitem{Venkatesh_2021_ICCV}
Rahul Venkatesh, Tejan Karmali, Sarthak Sharma, Aurobrata Ghosh, R.~Venkatesh
  Babu, L\'aszl\'o~A. Jeni, and Maneesh Singh.
\newblock Deep implicit surface point prediction networks.
\newblock In {\em Proceedings of the IEEE/CVF International Conference on
  Computer Vision (ICCV)}, pages 12653--12662, October 2021.

\bibitem{venkatesh2020dude}
Rahul Venkatesh, Sarthak Sharma, Aurobrata Ghosh, Laszlo Jeni, and Maneesh
  Singh.
\newblock Dude: Deep unsigned distance embeddings for hi-fidelity
  representation of complex 3d surfaces.
\newblock {\em arXiv preprint arXiv:2011.02570}, 2020.

\bibitem{pixel2mesh}
Nanyang Wang, Yinda Zhang, Zhuwen Li, Yanwei Fu, Hang Yu, Wei Liu, Xiangyang
  Xue, and Yu-Gang Jiang.
\newblock Pixel2mesh: 3d mesh model generation via image guided deformation.
\newblock {\em IEEE Transactions on Pattern Analysis and Machine Intelligence},
  43(10):3600--3613, 2021.

\bibitem{wang2021neus}
Peng Wang, Lingjie Liu, Yuan Liu, Christian Theobalt, Taku Komura, and Wenping
  Wang.
\newblock Neus: Learning neural implicit surfaces by volume rendering for
  multi-view reconstruction.
\newblock {\em NeurIPS}, 2021.

\bibitem{wang2022hfneus}
Yiqun Wang, Ivan Skorokhodov, and Peter Wonka.
\newblock Hf-neus: Improved surface reconstruction using high-frequency
  details.
\newblock {\em arXiv preprint arXiv:2206.07850}, 2022.

\bibitem{pixel2mesh++}
Chao Wen, Yinda Zhang, Zhuwen Li, and Yanwei Fu.
\newblock Pixel2mesh++: Multi-view 3d mesh generation via deformation.
\newblock In {\em ICCV}, 2019.

\bibitem{point-cloud-utils}
Francis Williams.
\newblock Point cloud utils, 2022.
\newblock https://www.github.com/fwilliams/point-cloud-utils.

\bibitem{disn}
Qiangeng Xu, Weiyue Wang, Duygu Ceylan, Radomir Mech, and Ulrich Neumann.
\newblock Disn: Deep implicit surface network for high-quality single-view 3d
  reconstruction.
\newblock In H. Wallach, H. Larochelle, A. Beygelzimer, F. d\textquotesingle
  Alch\'{e}-Buc, E. Fox, and R. Garnett, editors, {\em Advances in Neural
  Information Processing Systems}, volume~32. Curran Associates, Inc., 2019.

\bibitem{FoldingNet}
Yaoqing Yang, Chen Feng, Yiru Shen, and Dong Tian.
\newblock Foldingnet: Point cloud auto-encoder via deep grid deformation.
\newblock In {\em 2018 IEEE/CVF Conference on Computer Vision and Pattern
  Recognition}, pages 206--215, 2018.

\bibitem{yariv2021volume}
Lior Yariv, Jiatao Gu, Yoni Kasten, and Yaron Lipman.
\newblock Volume rendering of neural implicit surfaces.
\newblock In {\em Thirty-Fifth Conference on Neural Information Processing
  Systems}, 2021.

\bibitem{yariv2020idr}
Lior Yariv, Yoni Kasten, Dror Moran, Meirav Galun, Matan Atzmon, Basri Ronen,
  and Yaron Lipman.
\newblock Multiview neural surface reconstruction by disentangling geometry and
  appearance.
\newblock {\em Advances in Neural Information Processing Systems},
  33:2492--2502, 2020.

\bibitem{Ye_2022_CVPR}
Jianglong Ye, Yuntao Chen, Naiyan Wang, and Xiaolong Wang.
\newblock Gifs: Neural implicit function for general shape representation.
\newblock In {\em Proceedings of the IEEE/CVF Conference on Computer Vision and
  Pattern Recognition (CVPR)}, pages 12829--12839, June 2022.

\end{thebibliography}
}

\end{document}